\documentclass[acmlarge]{acmart}
\makeatletter
\newcommand{\confshort}{\acmConference@shortname}
\newcommand{\conffull}{\acmConference@name}
\newcommand{\confdate}{\acmConference@date}
\newcommand{\confloc}{\acmConference@venue}
\AtBeginDocument{
  \fancypagestyle{firstpagestyle}{
    \fancyhead{}%
    \fancyfoot[C]{}%
  }
  \fancyhf{}
  \fancyhead[LO]{\@headfootfont\shorttitle}%
  \fancyhead[RE]{\@headfootfont\@shortauthors}%
  \fancyhead[LE]{\@headfootfont\footnotesize \confshort, \confdate, \confloc}%
  \fancyhead[RO]{\@headfootfont\footnotesize \confshort, \confdate, \confloc}%
  \fancyfoot[C]{}%
}
\makeatother
\acmBooktitle{\conffull\@ (\confshort), \confdate, \confloc}
\AtBeginDocument{%
  }

\copyrightyear{2026}
\acmYear{2026}
\setcopyright{cc}
\setcctype{by}
\acmConference[FAccT '26]{The 2026 ACM Conference on Fairness, Accountability, and Transparency}{June 25--28, 2026}{Montreal, QC, Canada}
\acmBooktitle{The 2026 ACM Conference on Fairness, Accountability, and Transparency (FAccT '26), June 25--28, 2026, Montreal, QC, Canada}
\acmDOI{10.1145/3805689.3812226}
\acmISBN{979-8-4007-2596-8/2026/06}

\acmSubmissionID{1111}

\newcommand{\todo}[1]{{\color{red} #1}}

\usepackage{soul}
\usepackage{xcolor}
\soulregister\cite7 
\soulregister\ref7
\soulregister\todo7
\soulregister\begin7



\definecolor{revision_color}{HTML}{ACE8FF}
\definecolor{bad_color}{HTML}{FF0000}

\definecolor{fixed_color}{HTML}{B4FF94}
\newcommand{\fixed}[1]{#1}

\begin{document}

\title[CoAX: Cognitive-Oriented Attribution eXplanation User Model of\\ Human Understanding of AI Explanations]{CoAX: Cognitive-Oriented Attribution eXplanation User Model of Human Understanding of AI Explanations}

\author{Louth Bin Rawshan}
\email{louth@u.nus.edu}
\orcid{0009-0002-3272-8576}
\affiliation{%
  \institution{National University of Singapore}
  \country{Singapore}
}

\author{Zhuoyu Wang}
\email{wang.zhuoyu@u.nus.edu}
\orcid{0009-0009-8846-4750}
\affiliation{%
  \institution{National University of Singapore}
  \country{Singapore}
}

\author{Brian Y. Lim}
\email{brianlim@nus.edu.sg}
\affiliation{%
  \institution{National University of Singapore}
  \country{Singapore}
}
\authornote{Corresponding author}

\renewcommand{\shortauthors}{Rawshan and Lim}

\begin{abstract}
Explainable AI (XAI) aims to improve user understanding and decisions when using AI models.
However, despite innovations in XAI, recent user evaluations reveal that this goal remains elusive. Understanding human cognition can help explain why users struggle to effectively use AI explanations.
Focusing on reasoning on structured (tabular) data, we examined various reasoning strategies for different XAI methods (none, feature importance, feature attribution) in the decision task of anticipating AI decisions (i.e., forward simulation).
We i) elicited reasoning strategies from a formative user study, and ii) collected decisions from a summative user study.
Using cognitive modeling, we implemented the processes underlying each reasoning strategy and evaluated their alignment with human decision-making.
We found that our models better fit human decisions than baseline machine learning proxies, providing insights into which reasoning strategies are (in)effective.
We then demonstrate how the fitted model can be used to form hypotheses and investigate research questions that are costly to study with real human participants.
This work contributes to debugging human understanding of XAI, informing the future development of more usable and interpretable AI explanations.
\end{abstract}




\maketitle

\section{Introduction}

As AI pervades across domains, such as industrial quality assurance~\cite{zhang2021quality_ai}, sustainability~\cite{vinuesa2020ai_sustainability}, and socio-economic analysis~\cite{hardt2016equality}, 
explainable AI (XAI) has become crucial for users to understand model predictions~\cite{adadi2018peeking,abdul2018trends,wang2019designing}.
Despite the large number of XAI techniques proposed, their interpretability by users remains challenging and often ineffective~\cite{arora2022explain,bell2022s,poursabzi2021manipulating}, even when the methods satisfy desirable theoretical properties~\cite{rong2023towards,van2021evaluating} such as robustness or sparsity.
%
%
Arora et al. found that local explanations negatively affected decisions, but did not identify causes for this failure~\cite{arora2022explain}.
Poursabzi et al. found surprising results that explanations can hamper users' ability to detect model mistakes, and that XAI was ineffective despite multiple high-powered experiments~\cite{poursabzi2021manipulating}.
Hase and Bansal speculated that ineffective explanations had poor quality or users could not remember all instances, but these reasons were not validated~\cite{hase2020evaluating}.
Bell et al. loosely claimed participant "confusion" to describe SHAP explanations negatively affecting user understanding~\cite{bell2022s}. 
In general, the reasons posited lack rigor or involve conjecture, hindering insights to improve XAI interpretation.

Instead, we argue that explicit debugging of user reasoning would help identify proper XAI usage~\cite{wang2019designing}.
Inspired by works in cognitive architectures to improve UI design~\cite{luo2005predicting,yuan2021cogtool+} and for cognitive tutoring~\cite{anderson2013role}, we hypothesize that cognitive modeling to formally characterize human decisions with XAI can provide operational insight to improve the design of XAI or the education of their usage.
%
We propose CoAX, a cognitive model that integrates the theories of instance-based learning and cognitive boundedness to model the cognitive processes involved in interpreting XAI. We 
i) conducted a formative user study\footnote{User studies approved by university institutional review board.} to gain insights into how participants interpreted and used the XAI, 
ii) implemented the reasoning strategies observed, 
iii) collected decision data using a summative user study, and 
iv) validated CoAX against the observed behavioral data. 
%
%
We focus on tabular data, where features are structured and semantically meaningful.
We study Feature Attribution and the simpler Feature Importance explanations, which are widely adopted in user studies of XAI research, particularly SHAP and LIME~\cite{saarela2024recent, rong2023towards}.
We show that CoAX is able to simulate user decisions well, and also discuss further insights into the strategies preferred by users.
We further demonstrate CoAX's utility by modeling hypotheses with virtual experiments to investigate research questions that require higher cost and scale.
%
Our \textbf{contributions} are: 
1) user elicitation of reasoning strategies for attribution XAI,
2) user evaluation of their (in)effectiveness for AI decision tasks,
3) cognitive modeling to formally characterize (mis)interpretations.


\section{Related Work}


\subsection{Modeling User Performance on XAI}

Many studies have modeled human performance in XAI-assisted decision-making, but often with the intent of capturing behavioral trends rather than identifying cognitive processes.
A common approach is to train supervised ML models on user decision data to approximate human responses~\cite{virgolin2020formulaOfInterpretability,lage2018human, hilgard2021learningByHumansForHuman}. Beyond these proxy-based methods, some studies incorporate user feedback into the modeling process: 
Mozannar et al. identified AI error regions from users, revealing insights into how they understand AI outputs~\cite{mozannar2022teaching}, 
while Ma et al. trained a personalized decision tree on user decisions and allowed participants to refine it~\cite{ma2023should}. 
Conversely, Chen et al. trained a proxy only based on information from explanations, without human input or labels, showing some alignment with human decisions~\cite{chen2022use}.
%
%
Unlike the prior focus on surface-level user behaviors, we explicitly model user reasoning on XAI explanations. This cognitive-level modeling provides deeper insights into how users adopt certain strategies and how their reasoning evolves with XAI assistance. 

\subsection{Modeling Human Cognition}




Cognitive modeling has been applied across domains such as robotics~\cite{baron1994exploring}, autonomous driving~\cite{kolekar2021behavior,choi2021drogon,bhattacharyya2022modeling}, and agent-based simulations~\cite{dobson2019integrating,kennedy2011modelling} to predict and replicate human decision-making.
Many approaches rely on imitation learning~\cite{le2022survey} and machine theory of mind~\cite{rabinowitz2018machine}, which use large datasets to approximate human behavior but do not explicitly model cognitive mechanisms.
Reinforcement learning~\cite{sutton2018reinforcement,li2017deep} simulates learning through experience, but lacks representations of human memory and attention constraints~\cite{fuchs2023modeling}. 

Prior work in AI-assisted decision-making has formalized human reliance through stochastic and decision-theoretic models, such as Bayesian utility-maximizing acceptance policies~\cite{wang2022will}, Markovian trust dynamics~\cite{li2024utilizing}, and user heterogeneity~\cite{lu2024mix}. However, these behavioral abstractions are not cognitive models and do not examine XAI use.
Specific to XAI, 
Yang et al.~\cite{yang2022psychological} modeled saliency expectations via similarity, 
Labarta et al.~\cite{labarta2025cue} examined the relationship between legibility, readability, and user understanding, and 
Guo et al.~\cite{guo2024decision} established normative reliance benchmarks via rational agents.
Recently, LLMs have also served as proxies to estimate subjective helpfulness~\cite{de2024evaluating}.
However, these frameworks typically target observable outcomes or subjective perceptions rather than the cognitive processes and procedural reasoning strategies (i.e., mental algorithms) that users employ to interpret explanations.

Instead, we draw inspiration from instance-based learning (IBL)~\cite{logan1988toward,gonzalez2011instance}, which models decision-making as retrieving and adapting past experiences. Additionally, modeling cognitive limitations such as selective attention~\cite{anderson2004integrated} can help explain why users exhibit sub-optimal behavior, as demonstrated in models of typing behavior~\cite{shi2024crtypist} and pedestrian crossing decisions based on bounded optimal decision-making under noisy visual information~\cite{wang2024pedestrian}.
We apply these cognitive modeling techniques in a novel setting to analyze how users interpret or misuse XAI explanations. 
By integrating instance-based learning with cognitive constraints such as memory limitations and attentional biases, our approach can reveal how bounded cognition limits the usefulness of XAI.

\section{Overall Approach and Background}

\subsection{User-Centered Design for Cognitive Modeling}
\begin{figure*}[t]
    \centering
    \includegraphics[width=7.0cm]{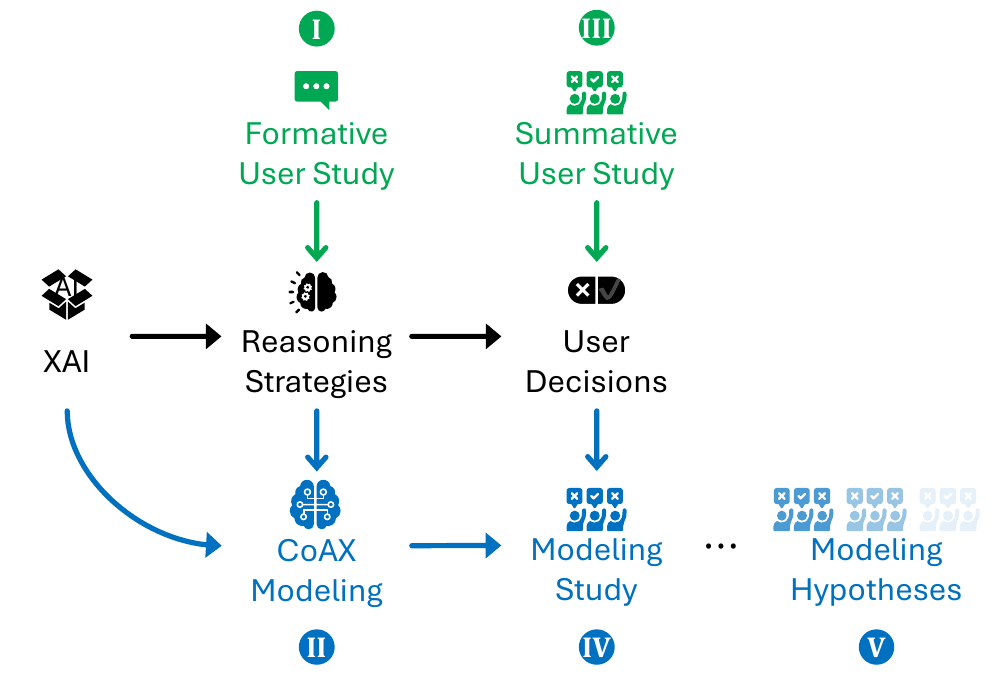}
    \vspace{-.25cm}
    \caption{
    Overall approach to modeling user behavior in XAI understanding using a five-step process: I) Formative study to elicit reasoning strategies, II) Implementing these strategies in the CoAX model, III) Summative study to obtain large-scale decision data, IV) Modeling study to analyze decision data using CoAX, and V) Testing hypotheses using CoAX modeling.
    }
    \label{fig:overall-approach}
\end{figure*}

We aim to simulate human decisions when learning from or using XAI by developing a cognitive model of the reasoning strategies.
This enables the diagnostic debugging of ineffective explanations---uncovering specific user or system deficiencies---while reducing the reliance on costly human-subjects experiments. 
To ensure cognitive fidelity,
we employ a user-centered design approach to determine the specifications for this model by conducting a formative study to elicit reasoning strategies, modeling these strategies using our framework, verifying them using a summative study, and then using the model to test hypotheses (Fig.~\ref{fig:overall-approach}).

\subsection{Forward Simulation Decision Task}


We prioritize evaluating user understanding of AI predictions as a necessary prerequisite to effective human-AI teaming~\cite{bansal2019beyond}, rather than focusing solely on downstream ground-truth decision outcomes.
Following Doshi-Velez and Kim~\cite{doshi2017towards}, like~\cite{chen2023understand}, we utilize \textit{forward simulation} tasks 
where users examine an instance's attributes alongside an explanation to estimate the AI's predicted output. 
This task evaluates understanding as a procedural capability foundational to goals such as model improvement, decision support, domain learning, and auditing~\cite{liao2022connecting, wang2019designing}. 
We elaborate on broader usage contexts in Section~\ref{sec:discussion_contexts} and the modeling extensions required for real-world decision tasks in Section~\ref{sec:discussion_realworld}.

Given an input $\mathbf{x} = (x_1, \dots, x_n)^\top$, the AI predicts
    $\hat{y} = f_{\text{AI}}(\mathbf{x})$,
and an XAI explanation model produces an explanation of this prediction:
    $\mathbf{e} = g_{\text{XAI}}(\mathbf{x}, \hat{y}, f_{\text{AI}})$.
The user's task is to estimate this AI prediction from the instance's attribute values with/without the AI explanation:
    $\Breve{y} = h_{\text{user}}(\mathbf{x}, \mathbf{e})$,
where ideally, $\Breve{y} \approx \hat{y}$, indicating good user understanding.
 

\subsection{Feature Importance and Attribution XAI}

To explain the AI prediction, we chose 
the popular XAI technique of feature attribution (e.g., LIME~\cite{ribeiro2016should}, SHAP~\cite{lundberg2017unified}).
These calculate signed \textit{feature attribution} $\mathbf{a}$ with each subscore $a_r$ (see Fig. \ref{fig:ui-xai}a,b,d) 
in support of or against the model prediction due to each feature $r$.
We also investigated the simpler \textit{feature importance} explanation which reports only non-negative supporting scores 
$\mathbf{a}^+ = \text{ReLU}(\mathbf{a})$
(see Fig. \ref{fig:ui-xai}a,b,c).

\subsection{Dataset and Application Context}

We examine how users interpret XAI in the common setting of tabular data classification, where the AI model provides predictions alongside feature-based explanations 
(feature importance or attribution).
Our study includes three widely used datasets with two labels (binary): 
Wine Quality~\cite{wine_quality_186}, Forest Cover Type~\cite{blackard1999comparative}, and Adult Income~\cite{kohavi1996scaling}.
Consistent with prior work~\cite{lage2019evaluation,dai2022counterfactual}, we selected these domains to reduce prior knowledge bias~\cite{lim2009and}, mitigating the need to evaluate with expert users, and providing domain-agnostic insights.
For generalization, for each dataset, we separately study one specific AI model (Multi-layer perceptron (MLP) or XGBoost~\cite{chen2016xgboost}) with one explanation method (LIME~\cite{ribeiro2016should}, SHAP~\cite{lundberg2017unified}); see mappings in  Table~\ref{tab:datasets-models} in the Appendix. 
We limit our study to cover only 5 attributes per dataset.



\section{Formative User Study of Reasoning Strategies on XAI} \label{sec:formative-user-study}




We conducted a user study to elicit human reasoning strategies when making decisions with XAI.
This formative study and the subsequent summative study were approved by the university institutional review board (IRB).

\subsection{Experiment Method and Procedure}
\begin{figure*}[t]
    \centering
    \includegraphics[width=10.6cm]{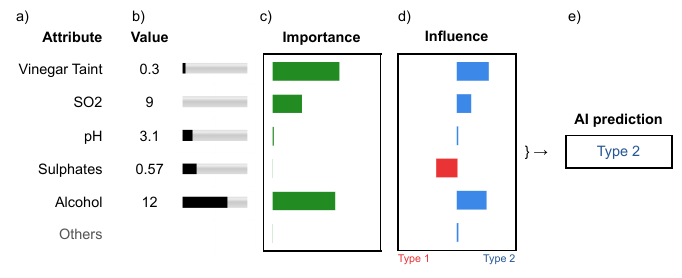}
    \vspace{-.5cm}
    \caption{UI components in the XAI interface.
    a) Attribute names of the instance being predicted.
    b) Value of each attribute with a bar to indicate how high/low.
    c) Importance scores of how important each attribute was for the current prediction.
    d) Attribution scores of importance and direction toward/against the prediction.
    e) Prediction label of AI model (anonymized to avoid any negative connotations, e.g., low-quality wine).
    a--b always shown; none, c or d shown depending on XAI type; e not shown for decision testing.
    }
    \label{fig:ui-xai}
\end{figure*}

We investigated the XAI condition (levels: None, feature Importance XAI, or feature Attribution XAI, where None means no XAI was given) with a partially counterbalanced \textit{within-subjects} experiment design, where each participant only saw 2 out of 3 versions across sessions to manage experiment length and participant fatigue.
Explanations were presented as in Fig.~\ref{fig:ui-xai}.
After a briefing on the study and UI, the participant had two sessions with 10 trials each. Each session 
covers only one XAI type to limit confusion. 
Like in Bo et al.~\cite{bo2024incremental}, for each trial, the participant
1) performs forward simulation without XAI (Fig.~\ref{fig:survey-training-no-xai}),
then 2) with XAI for the Importance and Attribution conditions only (Figs.~\ref{fig:survey-training-w-importance-xai},~\ref{fig:survey-training-w-attribution-xai}),
and 3) reviews the AI prediction feedback
(Figs.~\ref{fig:survey-training-feedback-no-xai},
\ref{fig:survey-training-feedback-w-importance-xai},
~\ref{fig:survey-training-feedback-w-attribution-xai}). 
We omitted showing None in session 2 to avoid carryover learning effects of seeing explanations in session 1.
See Appendix Fig.~\ref{fig:formative-experimental-design} for the procedure pipeline.

The participant is asked to think aloud throughout to articulate their reasoning. 
With consent, we record participant UI interactions and audio utterances for subsequent analysis. 
We thematically coded the transcripts of the think-aloud study using grounded theory analysis~\cite{charmaz2014constructing, strauss1990basics, glaser2017discovery}. 
First, the first and second co-authors conducted open coding with codes generated directly from participant utterances.
Next, these authors performed axial coding to identify relationships among the open codes and organized them into coherent reasoning strategies.
These two stages were conducted through multiple rounds of iteration.
Then, selective coding was used to merge similar strategies, including those observed across with- and without-XAI trials, but separated by XAI type. 
Finally, the identified reasoning strategies were reviewed by the last senior co-author and refined.
Throughout this process, we followed the grounded-theory heuristic of constant comparison to ensure that important distinctions between strategies were not overlooked.
%


\subsection{Findings: XAI Reasoning Strategies}
We recruited 20 participants from a local university: average age 23, 13 female, 
all undergraduate students from diverse majors. 
Since we aim to avoid prior knowledge bias, these lay users are suitable.
They were compensated \$14.76 USD in local currency for 60 min of their time. 
%
%
When shown XAI, participants used one common strategy for None, three for Importance XAI, and one for Attribution XAI. 
When not shown XAI, they first recalled or imputed relevant details before employing the same strategies.
We discuss details next.


\subsubsection{None XAI}
Despite the low number of attributes (five), participants focused on yet a smaller subset.

\textbf{Sensitive-features categorization.}
Participants tried to learn which attributes affected the model prediction in general, focusing on those, ignoring others. 
For example, Participant P10 was \textit{``assuming Vinegar Taint has something to do with predicting whether alcohol is high or low quality... I'm ignoring the rest.''} 
P3 learned which features had less influence on the output by noting \textit{``last time it was low quality, and pH was low, so I think pH is not that important.''}

\subsubsection{Importance XAI}
Participants exhibited three common reasoning patterns. Some made more use of the explanatory information and others focused on values based on explanation hints.

\textbf{Salient-features categorization.}
This involved focusing on attributes with the largest importance scores (salient) for that instance and recalling instances with similar values for those attributes. 
P14 recalled that \textit{``the Alcohol and the Sulphates are the most important [attributes]. I think it is Label 2, because I remember similar values [for those attributes] for one of the examples''}. P9 recalled a case that was \textit{``maybe similar to the last one [example], these two parameters [Alcohol and Sulphate] are high, ..., so I think Label 2.''}

Having learned this strategy, when predicting an AI decision without XAI, some participants imputed associations between important attributes and the predictions. 
The attributes were \emph{instance-dependent} in contrast to the participants in None who focused on a fixed subset of the attributes. 
P7 focused on two attributes where \textit{``based on the previous cases [...], when the Alcohol and the Sulphates were high, [the prediction] was Label 2''}, while stating for another instance that \textit{``when Vinegar Taint was high and Alcohol low, I think it [previous case] was type 1.''}

\textbf{Attribution sum.}
Like the previous strategy, participants attended to the salient attributes, 
but also individually assigned each an influence direction (either to Label 1 or 2) based on previous experiences, and summed up the signed importances. 
P12 thought \textit{``low Sulphates is Label 1 and high Sulphates is Label 2. But [I am choosing] Label 1, because I think if I add up the rest [of the importance scores] it outweighs [the effect of the high Sulphates].''}

Without XAI, participants estimated the importance and influence direction (Label 1 or 2) for each attribute, i.e., the attribution, and mentally summed them. 
P12 imputed that \textit{``Vinegar low [influences towards] Label 2, but pH low [influences towards] Label 1, Sulphates high is Label 2 [...].''}
P5 inferred \textit{``for this case, I will still predict as high quality, so even though the pH value is still high, I think that pH alone might not be enough to influence the overall results [i.e., the prediction].''}

\textbf{Importance categorization.}
Some participants learned patterns about the importance scores alone,
ignoring attribute values. 
This is flawed as an attribute's importance also depends on its value; relying only on importance patterns will underfit the true pattern. 
P6 noted that \textit{``if Vinegar Taint is more important than Alcohol then I think it is type 1, and otherwise it is type 2.''}
We did not observe this strategy when participants did not view XAI.


\subsubsection{Attribution XAI}
All participants converged to the same reasoning strategy.

\textbf{Attribution sum.}
Participants learned that a larger attribution sum towards a certain type would lead to the AI predicting that type. 
They often focused only on the largest attribution bars, but would consider the smaller bars for ambiguous cases with sum close to 0. 
P4 described this preference for the largest bars: \textit{``choosing Label 1 because Alcohol has a very large red bar...''}.
P16 approximated with counting: \textit{``there is more blue [Label 1] influence, so I will choose Label 1''}.
Without XAI, participants tried to recall attributions for predictions.
%
P4 recalled \textit{``For this, it [pH] was 3.9 and it was very negative, but here [around 50\%], it is not very negative.''}

\subsubsection{Reverting to None XAI}
Due to confusion, some participants ignored the explanation provided, resulting in falling back to the Sensitive-feature categorization strategy like with None XAI. 
For example, P10 \textit{``[thought] I used it [the importance explanation] a lot to see which values were important, but towards the end of it I felt it didn't really make sense because when I tried to do it based on the importance levels, sometimes it was wrong anyway.''}

\section{CoAX Cognitive Model}
We describe our cognitive model designed to approximate human behavior. We describe a base-model using the theory of instance-based learning, and specific extensions for each reasoning strategy from the formative study.



\subsection{Instance-Based Learning} \label{sec:instance-based-learning}

Our approach models user decision-making as a process of retrieving past exemplars from memory and comparing them to the current instance. This draws on Instance-Based Learning (IBL) Theory~\cite{gonzalez2011instance}, a comprehensive account of how humans make decisions from experience in dynamic tasks, which has been used to develop models that explain and predict behavior across categorization, judgment, and real-time interactions.

Retrieval is influenced by similarity, forgetting, and a retrieval threshold~\cite{anderson2004integrated,nguyen2023SpeedyIBL}. Exemplars with higher similarity and more recent storage are more likely to be retrieved.
Similarity between instance $\mathbf{x}$ and exemplar $\mathbf{x'}$ is defined as
\begin{equation}
    S(\mathbf{x}, \mathbf{x'}) = e^{-\alpha d(\mathbf{x}, \mathbf{x'}) + A},
    \label{eq:exemplar-similarity}
\end{equation}
where \( d \) is the normalized Euclidean distance, \( \alpha \) is distance sensitivity, and \( A \) is the activation of exemplar $\mathbf{x'}$. The activation of each exemplar in memory follows
$
    A = -\lambda \ln (\Delta t),
    \label{eq:memory-activation}
$
where \( \lambda \) is the memory decay rate and \( \Delta t \) is the elapsed time since storage~\cite{anderson1991reflections}. Only exemplars with \( A \geq \rho \) are retrieved, where $\rho$ is the memory retrieval threshold.
To predict the label of an instance $\mathbf{x}$, exemplars are retrieved and aggregated to determine a probability distribution over labels, weighted by similarity, based on the Generalized Context Model~\cite{nosofsky1986attention}:
\begin{equation}
    P(y | \mathbf{x}) = 
    \left.
    {\sum\nolimits_{\mathbf{x'}: y_{\mathbf{x'}} = y} S(\mathbf{x}, \mathbf{x'})}
    \middle/
    {\sum\nolimits_{\mathbf{x'}} S(\mathbf{x}, \mathbf{x'})}.
    \right.
\end{equation}

\subsection{Reasoning Strategy Models}

We describe how each of the reasoning strategies observed during the formative study is implemented by extending the base IBL model,
by specifying which parts of the instance and explanation are stored as exemplars in memory and how these are used to make predictions after recall.
We keep the model names consistent with the strategy names for clarity.

\textbf{Sensitive-features categorization} selects a fixed subset of \( k \) attributes that best differentiate between categories, modeling participants who focused on attributes that \textit{``have something to do with predicting''}, while \textit{``ignoring the rest''} (P10).
Features are ranked by their discriminative power, measured using the $t$ statistic:
\begin{equation}
    t_r = \left.
    | \mu_1^r - \mu_2^r |
    \middle/
    {\sqrt{{s_1^{r\,2}}/{N_1} + {s_2^{r\,2}}/{N_2}}},
    \right.
\end{equation}
where \( \mu_1^r \) and \( \mu_2^r \) are the mean values of the $r$th feature \( x_r \) for stored exemplars in each label group, \( s_1^r \) and \( s_2^r \) are their standard deviations,
and \( N_1 \), \( N_2 \) are the numbers of exemplars for classes 1 and 2, respectively. 
Features are ranked by \( t_r \), with higher values indicating more discriminative features, to select the top \( k \) features which are then evaluated for similarity as described in Section~\ref{sec:instance-based-learning}. 

\textbf{Salient-features categorization} dynamically selects the top-\( k \) features $\mathcal{F}_k$ with the highest importance scores $\mathbf{a}^+$, 
modeling participants who focused on \textit{``the most important [attributes]''} and recalled cases with \textit{``similar values [for those attributes]''} (P14).
Distance is based only on these features:  
\begin{equation}
    d(\mathbf{x}, \mathbf{x'}) = \frac{1}{k}\sum\nolimits_{r \in \mathcal{F}_k} (x_r - x'_r)^2.
    \label{eq:salient-features-categorization}
\end{equation}
Only these features of the exemplar are stored in memory, such that even without XAI, similarity $S$ between a new instance and this exemplar is calculated based only on these features.

\textbf{Attribution sum} computes a single log-odds score by summing attributions across features,
modeling participants who assessed the accumulation of \textit{``more blue [Label 1] influence''} (P16) to weigh their decision on the AI prediction.
In the cases of Importance XAI or testing without XAI, for each feature \(r\), the model retrieves exemplars with similar values from memory to estimate a per-feature label \(y_r \in \{-1, +1\}\). 
Using each feature importance score $a_r^+$, each feature has a weighted vote for the label, which we normalize with the sigmoid function into a probability:
\begin{equation}
    P(y=1) = \sigma( \zeta \sum\nolimits_{r} a_r^+ y_r ),
    \label{eq:attribution-sum}
\end{equation}
where \(\zeta\) is the feature-class sensitivity parameter.
When Attribution XAI is provided, $y_r$ is provided, so participants can directly sum all attribution values to make a prediction.

\textbf{Importance categorization} 
uses the same method as Sensitive-features categorization, but treats importance scores $\mathbf{a}^+$ as feature values instead of using the actual values $\mathbf{x}$, modeling participants who inferred directly from importance patterns, e.g., \textit{``if Vinegar Taint is more important than Alcohol then I think it is type 1''} (P6).

\textbf{Random}, $P(y=1)=0.5$, accounts for participants who may be non-conscientious and decide randomly.

\subsection{Model Parameters}
In total, the CoAX model has several parameters that need to be estimated for each virtual human:
1) exemplar distance sensitivity $\alpha$ in Eq.~\ref{eq:exemplar-similarity},
2) retrieval activation threshold $\rho$,
3) number of features $k$ selected via the sensitive-features or salient-features strategies to compare exemplars, and 
4) the feature-class sensitivity \(\zeta\) in Eq.~\ref{eq:attribution-sum} for the attribution sum strategy.
Memory decay rate $\lambda$ in Eq.~\ref{eq:memory-activation} is set as 0.5 for all virtual humans based on prior works~\cite{anderson2004integrated}. 
The parameters are the same for each virtual human for all datasets.

\section{Summative Study of XAI Understanding} \label{sec:summative-user-study}
We conducted a summative user study to 
evaluate the usefulness of XAI type for user understanding.
This also provides a dataset to validate our CoAX cognitive model.

\subsection{Experimental Design}

\begin{figure}[t]
  \centering
  \includegraphics[width=0.85\textwidth]{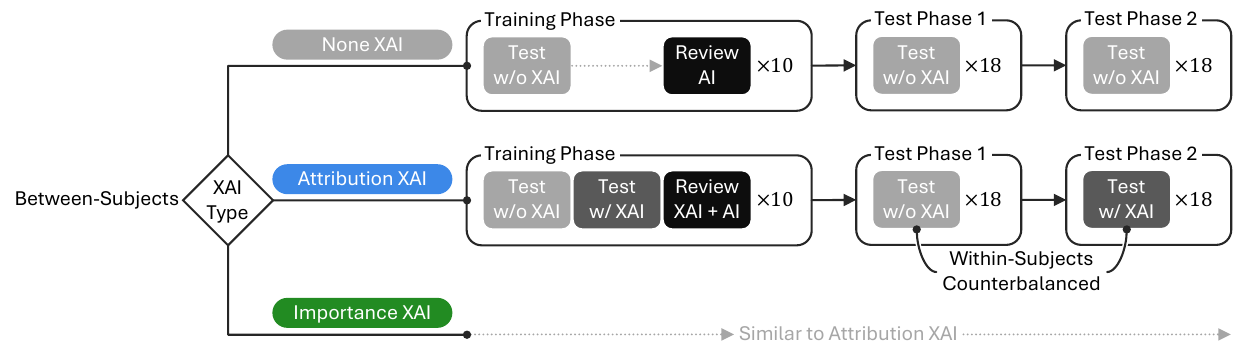}  
  \vspace{-.15cm}
    \caption{Experiment pipeline of each session in the summative user study. Similar phases---10 training trials, 2 $\times$ 18 test trials \\(within-subjects counterbalanced)---across XAI types (between-subjects), but with different information tested \fixed{per} condition.}
    \label{fig:summative-experimental-design}
\end{figure}
We manipulated independent variables: 
XAI type (None, Importance, and Attribution; \textit{between-subjects} to avoid learning effects \fixed{across} XAI types), and
Test setting (with or without XAI; within-subjects).
We measured dependent variable: Human label (Label 1 or 2) of the participant's forward simulation of the AI decisions.




\subsubsection{Experiment Procedure} 
\label{sec:summative-user-study-procedure}
Participants were introduced to the study, gave consent, completed a tutorial on the XAI interface, and answered screening questions to confirm comprehension. Each then completed two sessions with a break between. 
Each session consisted of a training phase (10 trials with fixed sequence\footnote{This facilitates active learning through interleaving testing and learning across trials, by cognitively forcing engagement~\cite{buccinca2021trust} before revealing answers. Our pilot studies verified this benefit, in agreement with prior work \cite{bo2024incremental}.}: predict without XAI, then with XAI for Importance and Attribution, followed by feedback), and two testing phases (18 trials each, with XAI and without XAI, randomly ordered; w/ XAI = w/o XAI for the None condition) (Fig.~\ref{fig:summative-experimental-design}). 
The study ended with demographic questions. 



We recruited 335 participants (199 female, median age 29) via Prolific. An additional 165 participants were excluded based on a straightforward comprehension screening (see Appendix Figs.~\ref{fig:survey-screening-importancexai},~\ref{fig:survey-screening-attributionxai} for questions); this inclusion rate is comparable to prior literature (e.g., 71\% in~\cite{swaroop2025personalising}
 and 76\% in~\cite{wang2023watch}). 
Participants were assigned to one of three XAI types (None $\times1$, Importance $\times4$, Attribution $\times2$), randomly weighted based on the number of strategies found per XAI type identified in the formative study\footnote{Reasoning Strategy is our key independent variable for analysis, especially with cognitive modeling (not XAI type), so we sought to balance them.
This also mitigates undersampling of the more diverse reasoning strategies in Importance XAI.}.
They were randomly assigned \textit{between-subjects} to one of three datasets (Wine Quality, Adult Income, Forest Cover Type), 
with the corresponding allocated XAI model described in Appendix Table~\ref{tab:datasets-models}.
There were no educational background prerequisites. 
Each was compensated £9.00 for an average completion time of 35 minutes. The variation in datasets and XAI types is intended to assess the generalizability of CoAX across settings and test whether lab-elicited strategies persist in a broader online population, rather than to compare conditions directly. 

\subsection{Results}
\label{sec:user-study-main-results}

\begin{figure*}[t]
  \centering
  \includegraphics[width=1.0\textwidth]{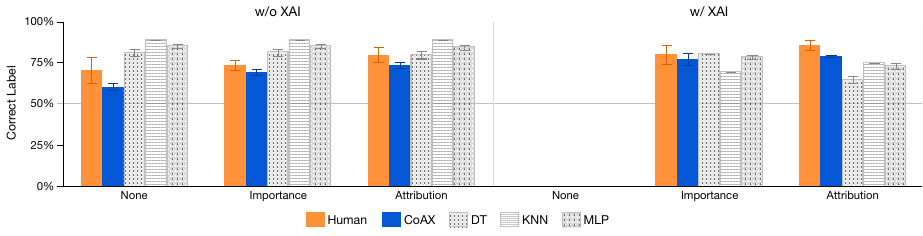}  
  \caption{Results of summative user study (orange) compared to virtual proxies (CoAX blue, ML-based grey) of label correctness by XAI Type for the Adult Income dataset.
    Error bars 95\% CI.}
  \label{fig:summative-xaitype}
\end{figure*}

 For analysis, we used the 72 test trials (2 sessions $\times$ (18 test w/o XAI + 18 test w/ XAI trials)) per participant and fit a linear mixed-effects model with label correctness as the response, XAI type, and testing condition (with vs. without XAI) as fixed effects, their interaction, and participant as a random effect.
We analyzed the results for each dataset separately due to their independence.
We performed a Tukey HSD test across the five (XAI type $\times$ Testing condition) combinations with $p < .05$ (detailed significance results in Appendix~\ref{sec:appendix-tukey-hsd-by-xaitype}).
Fig.~\ref{fig:summative-xaitype} (orange bars) shows participant performance on the forward simulation task (Label Correctness) for the Adult Income dataset.
See Appendix~\ref{sec:appendix-coax-vs-baselines} for results of other datasets.

We report general trends here, as specific significance patterns vary by dataset.
When tested without explanations (w/o XAI), participants trained with Attribution or Importance XAI significantly outperformed those who never received explanations (None), indicating sustained learning from prior exposure.
When explanations were available at test time (w/ XAI), Attribution XAI performed marginally better than Importance XAI and significantly better than None.

\section{Modeling Study of Human--CoAX Fit} \label{sec:modeling-analysis}

To understand why some explanations were more effective than others, we used the CoAX cognitive model to infer participants' reasoning strategies, as direct probing across all 92 trials (2 sessions $\times$ (10 training + 2 $\times$ 18 test w/o and w/)) was infeasible. Analyses were run on a laptop with an i5-12500H processor without using the GPU.



\subsection{Fitting CoAX Parameters to Each Participant} \label{sec:coax-individual-fit}

We fit one CoAX strategy model per session (18 trials) using Gaussian process-based Bayesian optimization~\cite{snoek2012practical}, and selected the strategy $\pi$ with the lowest BIC~\cite{vrieze2012model}. 
Refer to Appendix Table~\ref{tab:avg-mu-sd-hyperparams-by-strategy} for the distribution of the fitted parameters.
Table \ref{tab:proxy-model-fits} shows the goodness-of-fit of the CoAX model using the NLL and BIC metrics.
Next, we use the inferred reasoning strategies to analyze strategy prevalence and performance across XAI conditions.

 \begin{table*}[t]
    \centering
    \caption{Comparison of model fits across XAI types, averaged over all participants from the 3 datasets, using negative log-likelihood (NLL) and Bayesian information criterion (BIC). Mean $\pm$ SD across participants is reported. Best fit is bolded.}
    \resizebox{\linewidth}{!}{
    \begin{tabular}{l cccc cccc cccc}
        \toprule
        & \multicolumn{4}{c}{\textbf{None XAI}} & \multicolumn{4}{c}{\textbf{Importance XAI}} & \multicolumn{4}{c}{\textbf{Attribution XAI}} \\
        \cmidrule(lr){2-5} \cmidrule(lr){6-9} \cmidrule(lr){10-13}
        Metric & DT & KNN & MLP & CoAX & DT & KNN & MLP & CoAX & DT & KNN & MLP & CoAX \\
        \midrule
        NLL $\downarrow$
        & 0.61 $\pm$ 0.10 & 0.58 $\pm$ 0.14 & 0.61 $\pm$ 0.07 & \textbf{0.47 $\pm$ 0.22}
        & 0.60 $\pm$ 0.10 & 0.59 $\pm$ 0.11 & 0.60 $\pm$ 0.09 & \textbf{0.38 $\pm$ 0.23}
        & 0.58 $\pm$ 0.09 & 0.46 $\pm$ 0.16 & 0.43 $\pm$ 0.11 & \textbf{0.31 $\pm$ 0.21} \\
        BIC $\downarrow$
        & 27.7 $\pm$ 3.6 & 26.7 $\pm$ 5.0 & 27.7 $\pm$ 2.5 & \textbf{26.0 $\pm$ 7.9}
        & 27.4 $\pm$ 3.6 & 27.0 $\pm$ 4.0 & 27.4 $\pm$ 3.2 & \textbf{25.8 $\pm$ 7.2}
        & 26.7 $\pm$ 3.2 & 22.3 $\pm$ 5.8 & 21.3 $\pm$ 4.0 & \textbf{25.7 $\pm$ 7.6} \\
        \bottomrule
    \end{tabular}}
    \label{tab:proxy-model-fits}
\end{table*}

\subsubsection{Prevalence of Reasoning Strategies}
\begin{figure*}[t]
  \centering
  \includegraphics[width=12.5cm]{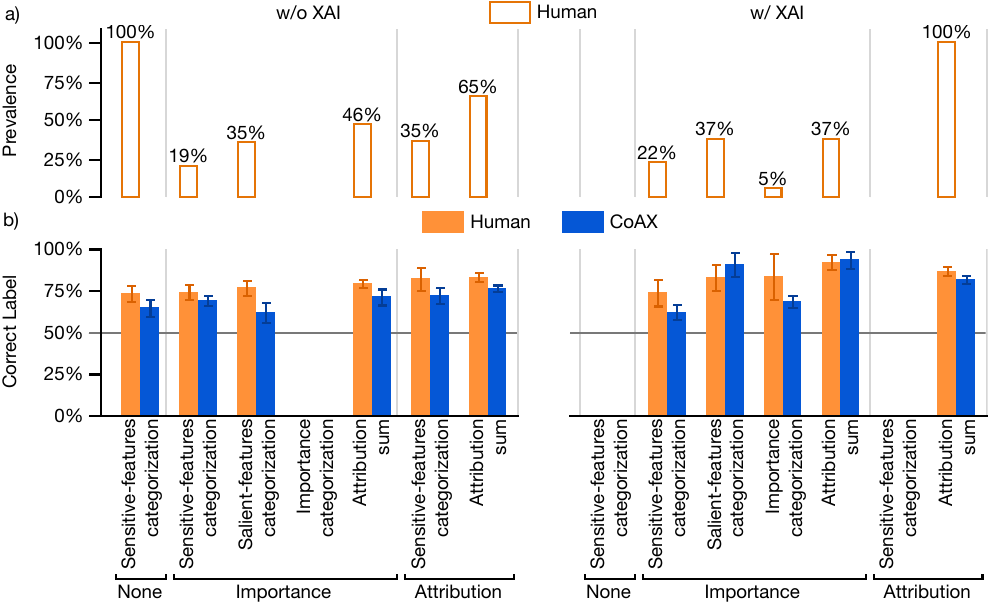}  
  \vspace{-0.2cm}
  \caption{Results of summative user study (orange) and CoAX simulation (blue) of prevalence (a) and label correctness (b) by Reasoning Strategy for the Adult Income dataset. Pearson correlation between Human and CoAX, r = 0.952.
    Error bars indicate 95\% CI.}
    \label{fig:summative-xaitype-strategy}
\end{figure*}

Fig. \ref{fig:summative-xaitype-strategy}a shows the prevalence of participant sessions that were inferred to use each Reasoning Strategy for different conditions for the Adult Income dataset (see Appendix~\ref{sec:appendix-prevalence-strategy} for figs. for other datasets). 
Across datasets, in the Test w/ Attribution XAI and None conditions, participants were exclusively matched to a single strategy: Attribution sum and Sensitive-features categorization, respectively. Without Attribution XAI at test time, only 32--65\% retained the Attribution sum strategy; the rest reverted to Sensitive-features categorization.
When testing with Importance XAI, Attribution sum (37--39\%) was most popular, followed by Sensitive-features categorization (22--35\%) and Salient-features categorization (18--37\%), with Importance categorization (3--7\%) having very sparse usage.
The significant usage of Sensitive-features categorization, which
ignores the explanation provided, suggests participants \textit{under-rely} on the XAI.
Interestingly, Salient-features categorization was more popular than Sensitive-features categorization for the Adult Income dataset. 
This could be due to participants being familiar with the domain, enabling \fixed{them} to contextually focus on different attributes.

14.2\% of participants were best fit to the Random strategy\footnote{Our analysis of Test Sessions 1 and 2 did not find a difference in the number of participants with this weak reasoning strategy (14.3\% vs. 14.0\%), suggesting no strategy shift or increasing disengagement among participants.}. 
This proportion is consistent with prior work~\cite{maddox2005delayed, freedberg2017comparing}.
Focusing on known, effortful reasoning strategies, we excluded these participants from strategy-level analyses in the following section.
However, they remain in objective comparisons against baseline models to ensure fairness.

\begin{table}[t]
  \centering
  \caption{
    Statistical results from a Tukey HSD test at $\alpha = 0.05$ showing reasoning strategies that have similar correctness rates (same group letters: A, B, C, etc.) of Human responses aligned with CoAX simulation for the forward simulation task on the Adult Income dataset.
    The same results are visualized in Fig.~\ref{fig:summative-xaitype-strategy}.
    Results for other datasets in Appendix~\ref{sec:appendix-tukey-hsd-strategies}.
  }
  \vspace{-0.1cm}
  \includegraphics[width=12.4cm]{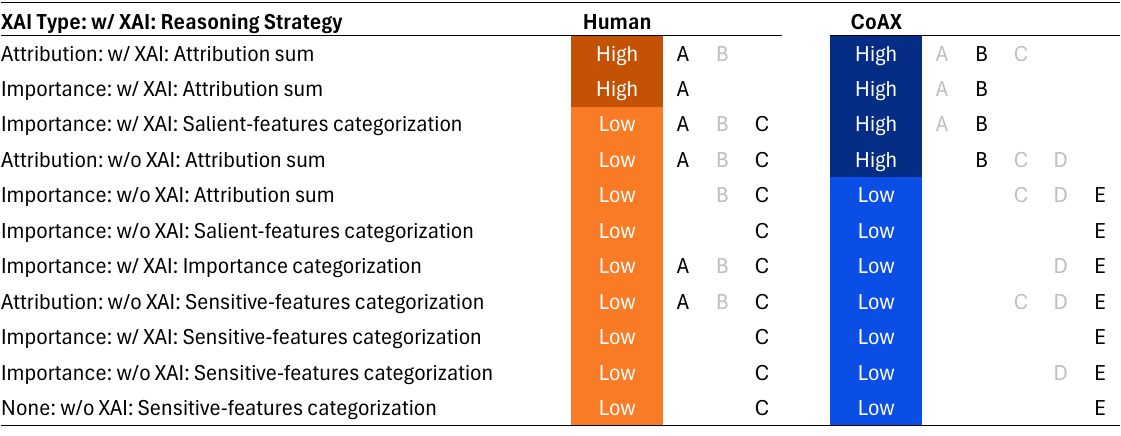}
  \label{tab:tukey-hsd-adult-income-strategies}
\end{table}

\subsubsection{Efficacy of Reasoning Strategies}

We analyzed the efficacy of reasoning strategies based on human responses, and compared them with CoAX simulations to 
examine CoAX's faithfulness in representing human reasoning on XAI and decisions on AI.
We synthesized 25 virtual participants for each of the 11 conditions (XAI type: Testing with or without XAI: Reasoning Strategy) using the parameter distributions fitted to human data, which were held constant across datasets.  


Figure~\ref{fig:summative-xaitype-strategy}b shows human (orange) and CoAX (blue) performance on the Adult Income dataset, with close alignment across the 11 conditions.
To evaluate strategy efficacy, we conducted separate Tukey HSD tests comparing performance across the 11 conditions.  
Table~\ref{tab:tukey-hsd-adult-income-strategies} reports the resulting rankings for the Adult Income dataset, showing that CoAX matches the human ranking within margins of uncertainty.  

Across datasets, 
when tested with Importance XAI, Attribution sum achieved the highest performance, followed by Salient-features categorization, Importance categorization, and Sensitive-features categorization.
A similar ordering was observed when testing without explanations, except that Importance categorization was absent.
When participants were trained with Attribution XAI and tested without explanations, Attribution sum outperformed Sensitive-features categorization.  
The rankings of all 11 conditions are consistent across the Adult Income and Forest Cover datasets.
However, for the Wine Quality dataset, Salient-features and Importance categorization underperformed, perhaps 
due to the linear Attribution/Importance explanations being misleading.

\subsection{Comparing CoAX Against Other Machine Learning Decision Proxies}
\label{sec:compare-decision-proxy}
We evaluated how well CoAX aligns with human responses from our summative study, compared to alternative machine learning proxies that simulate user behavior.

\textbf{CoAX.}
We simulated another 50 virtual participants\footnote{Higher sample size for stronger statistical power.} for each XAI type, with the same prevalence of reasoning strategies as in Section \ref{sec:coax-individual-fit}
to measure the labeling correctness of all CoAX cognitive models.
Fig. \ref{fig:summative-xaitype} shows that, as a population, CoAX (blue) is aligned with the human (orange) results (Pearson correlation, $r = 98.8$\%).


\textbf{Machine Learning Proxies.}
We compared CoAX against baseline ML proxies: Decision Tree (DT), KNN, and MLP. 
For each session, the proxy model was trained on 10 training instances and tested on 18 test instances.
For each model family, two models were trained per session: one for testing without XAI (using only feature values $\mathbf{x}$), and one for testing with XAI (feature vector concatenated with feature importance $\mathbf{a}^+$ or attribution $\mathbf{a}$). 
Unlike traditional ML training, these models are trained on the AI prediction labels for the forward simulation task, with hyperparameter tuning on human labels to emulate each user, not the ground truth labels. 
Details of the hyperparameters tuned are in Appendix~\ref{sec:appendix-tuning-hyperparams}.

Table \ref{tab:proxy-model-fits} shows these ML proxies do not model individual decisions as closely as CoAX, while
Fig. \ref{fig:summative-xaitype} (grey patterned bars) shows this mismatch at a reasoning strategy level.
When tested without XAI, DT aligns well with human performance but MLP and KNN overperform, perhaps due to overfitting to AI predictions rather than human behavior.
With Attribution XAI,
all baselines underperform humans, perhaps because they cannot efficiently learn 
the Attribution sum process.

\subsection{Understanding due to Cognitive Parameters.}

\begin{figure}[t]
  \centering
  \includegraphics[width=13.5cm]{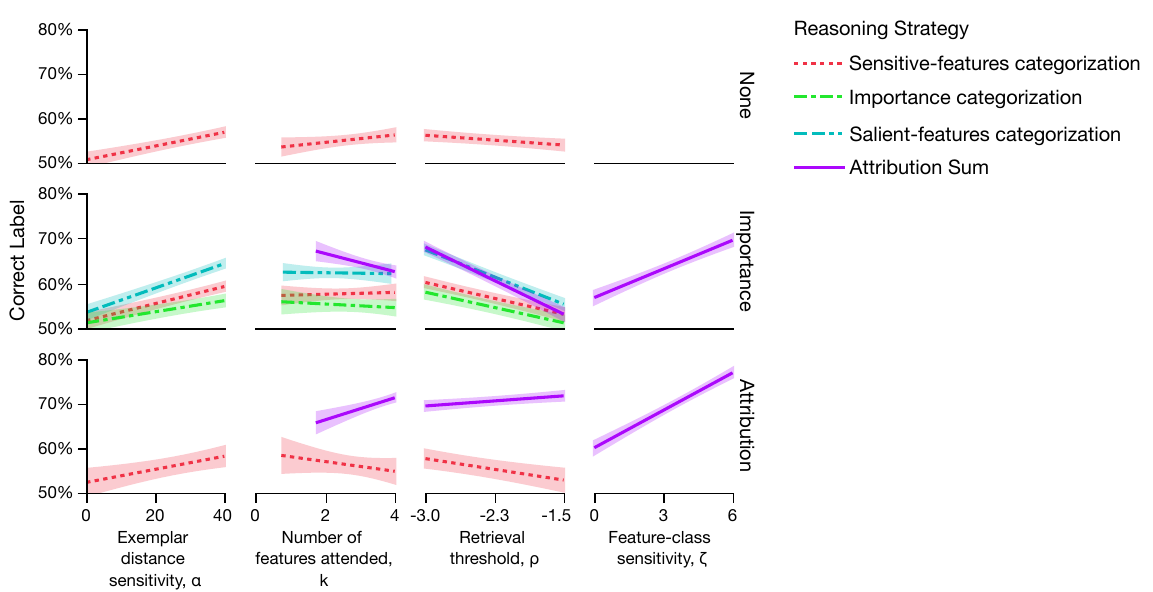}
  \vspace{-0.25cm}
  \caption{
  \fixed{Effect of} fitted CoAX cognitive parameters on label correctness by reasoning strategy, averaged for all datasets.
  }
  \label{fig:summative-params}
\end{figure}

We examined how the performance of the generated CoAX models vary with their cognitive parameters (Fig.~\ref{fig:summative-params}).
Across XAI types, label correctness increases with exemplar distance sensitivity $\alpha$ and feature-class sensitivity $\zeta$, as greater sensitivity favors more similar exemplars and reduces interference from irrelevant ones.
Correctness decreases with higher retrieval thresholds $\rho$, showing that increased forgetting leads to poorer performance.
No consistent trends are observed with respect to the number of features attended $\kappa$.

\section{Modeling Hypotheses of Summative User Studies}





Having validated CoAX against human reasoning on XAI (Section~\ref{sec:coax-individual-fit}), we use it to generate \textit{hypotheses} for research questions that are costly to study with real human participants, which future work could validate.
We examine how explanation understandability evolves by increasing user training and by varying XAI complexity and XAI models.
\subsection{Method}
For the experiment design, along with XAI Type (None, Importance, Attribution) and Testing conditions (w/, w/o XAI) as primary independent variables (IV1, IV2), we had secondary IVs: 
IV3) Number of Training Instances (1--13),
IV4) Number of Attributes (1--9), and
IV5) XAI model (SHAP~\cite{lundberg2017unified}, LIME~\cite{ribeiro2016should}, Integrated Gradients~\cite{sundararajan2017axiomatic}, Input Gradients~\cite{shrikumar2017learning}).
For each IV1$\times$IV2 level, we varied IVs 3--5 in separate experiments.
For generality, we varied Dataset (Wine Quality, Forest Cover) as the random variable.
As control variables, we set all AI Models as MLP, and XAI method = SHAP when varying IV3 and IV4.
We measured their impact on dependent variable (DV) Correctness of forward simulation by the virtual participant on the AI decisions.
Simulating 100 CoAX virtual participants per condition, we used the same distributions of cognitive parameters and reasoning strategies as in the Modeling Study (Section~\ref{sec:coax-individual-fit}).
Virtual participants did the same procedure as in Section~\ref{sec:summative-user-study-procedure}, subject to varying trials based on IV3.

\subsection{Results}

Figs.~\ref{fig:vary-num-training-and-attributes} show the results for the two virtual experiments on Wine Quality.
See Appendix Figs.~\ref{fig:vary-num-training-and-attributes-forest-cover} for Forest Cover.

\textit{Number of Training Instances.}
As expected, with more training instances, understanding Correctness increases and plateaus when testing w/o XAI, and understanding is better w/ XAI (Fig.~\ref{fig:vary-num-training-and-attributes}a). 
However, understanding Correctness w/ Attribution XAI is highest and already plateaued regardless of the number of instances.

\textit{Number of Attributes.}
Unlike with training instances, increasing attributes shown in the instance leads to lower understanding Correctness w/o XAI, perhaps due to information overload.
Nevertheless, this decrease is mitigated with Importance and increases with Attribution (Fig.~\ref{fig:vary-num-training-and-attributes}b).

\textit{XAI Models.}
All XAI models performed better for Attribution than Importance XAI (Fig.~\ref{fig:vary-num-training-and-attributes}c) and were better than None, though there were notable differences.
SHAP and LIME performed similarly, though LIME was less helpful for Attribution, perhaps due to its poorer AI faithfulness.
Input Gradients was the worst performing, lacking the completeness and robustness properties of Integrated Gradients~\cite{sundararajan2017axiomatic}.

\begin{figure}[t]
  \centering
  \includegraphics[width=14.8cm]{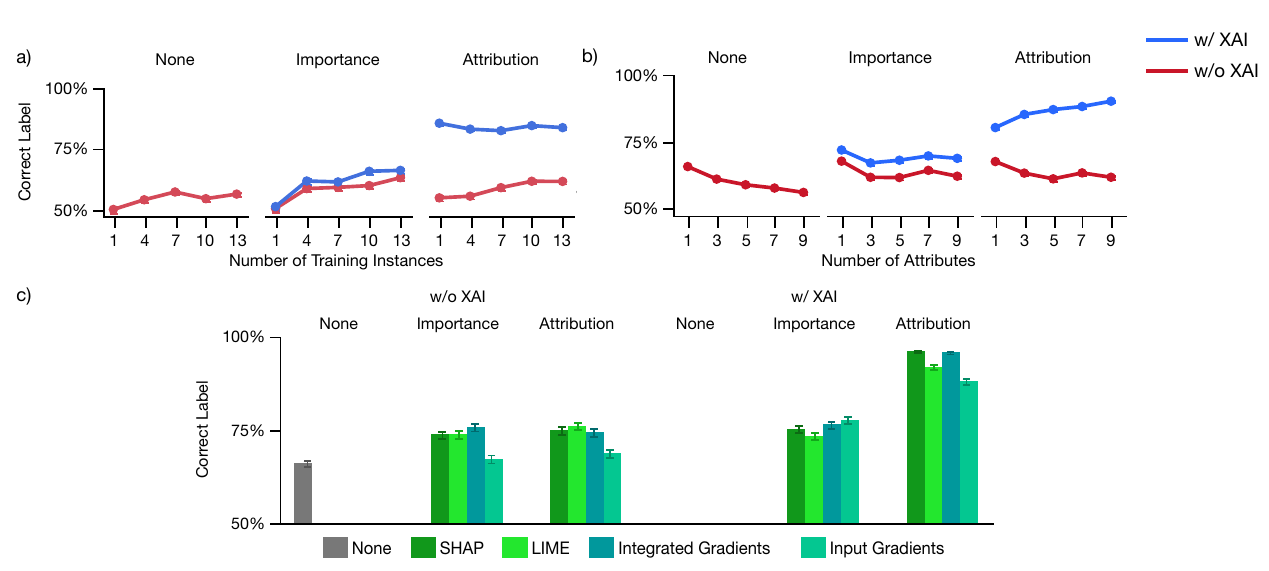}
  \vspace{-0.3cm}
  \caption{Modeling hypotheses study results of forward simulation correctness on the Wine Quality dataset, by XAI type, testing condition, a) number of training instances, b) number of attributes, and c) XAI model.}
  \label{fig:vary-num-training-and-attributes}
\end{figure}

\section{Discussion}








\subsection{Design Implications}

\subsubsection{Takeaway of Studies: How XAI Type Shapes Reasoning Strategies}

Our modeling with CoAX has found several reasoning strategies to help identify good and debug poor user understanding of XAI.
Specifically, we identified how explanation formats (None vs. Importance vs. Attribution) shaped reasoning strategies.
We found that using Attribution XAI primarily resulted in consistent arithmetic-like thinking (Attribution sum) where users approximately summed subscores in support of or against the prediction label and weighed opposing evidence, likely facilitated by the polar representation.
Conversely, using Importance XAI led to diverse reasoning strategies, including misuses, such as Importance categorization that wrongly interprets subscores as attribute values for pattern matching.
This stems from the lack of context due to the lack of contrastive explanations (negative attributions).
This suggests that reducing explanation contexts will lead to less consistent user reasoning, including more weak ones. 
Particularly interesting is the possibility that saliency methods~\cite{selvaraju2017grad} that clip negative attributions have diminished interpretability.

\subsubsection{Implications of CoAX Model for Design Recommendations}

CoAX can help debug which reasoning strategies lead to poor user understanding. But what does this imply for improving XAI?
We consider some approaches for various stakeholders:
HCI researchers could design tutorials 
to promote more reliable reasoning strategies and discourage common misinterpretations.
XAI UI designers could use CoAX to systematically \textit{select} explanation hyperparameters by modeling the trade-off between information gain and cognitive load (e.g., running ablation studies to identify the optimal feature count). 
Interaction designers could prompt (``cognitively force''~\cite{buccinca2021trust}) users to reflect on salient features for specific instances, encouraging Salient-features categorization and suppressing Sensitive-feature categorization.
XAI developers could \textit{optimize} XAI methods to improve user decision performance by regularizing for desiderata
hypothesizes as impacting \fixed{decision tasks~\cite{nofshin2024sim2real}},
or by training XAI models via backpropagation from CoAX-labeled decisions using a pre-trained neural network proxy~\cite{li2024utilizing}.
%

\subsubsection{Methodological Risks of Over-Reliance on CoAX}
CoAX is meant for preliminary hypothesis formation and testing, but real user studies must still be conducted for conclusive claims. 
Additionally, the reasoning strategies implemented in CoAX should be informed by formative user studies
of representative target users rather than solely conceived by XAI developers who may not think like the end-users. 
Finally, CoAX models forward simulation performance, which omits other decision tasks and desiderata, \fixed{like} counterfactual simulation performance, satisfaction and trust. 
Neglecting these goals risks biased and misaligned consequences, i.e., Goodhart's law~\cite{strathern1997improving}.

\subsubsection{Initial Scope, Other Contexts, and Sociotechnical Considerations}
\label{sec:discussion_contexts}

Our modeling study (Section~\ref{sec:compare-decision-proxy}) showed that the CoAX reasoning strategies fit human decisions better than machine learning proxies, 
suggesting the strategies capture a procedural logic rather than mere statistical correlation.
However, these strategies are an initial partial set of four derived from our formative study on forward simulation decisions as a proxy for understanding~\cite{doshi2017towards} of university students and online crowdworkers.
A fuller range of reasoning strategies could be elicited and modeled using our framework (Fig.~\ref{fig:overall-approach}) from more diverse users (e.g., data scientists, domain experts, elderly), and usage contexts or tasks (e.g., model improvement, decision support, domain learning, auditing)~\cite{liao2022connecting, wang2019designing}.
For example, although we did not observe the plausible Attribution Categorization reasoning strategy (for retrieving instances by similar attribution values), it could be revealed by increased task familiarity, expert domain knowledge, or shifting user goals~\cite{payne1993adaptive}.
%
Sociotechnically, these potential strategy divergences suggest that different user populations may interpret explanations differently. 
While CoAX currently underrepresents strategies from marginalized groups, it provides a pipeline to incorporate their specific reasoning strategies and cognitive parameters. 
Integrating these into the framework enables researchers to move beyond "one-size-fits-all" XAI toward systems optimized for varied reasoning styles and more equitable outcomes.

\subsection{Generalization}
%
Generalizing across three datasets, we investigated forward simulation for tabular data with binary labels and 5 attributes.
Future work can consider extensions to other explanation types, modalities, and decision tasks.

\subsubsection{XAI Types}
Beyond Feature Attribution, CoAX could be extended to consider other XAI types (Appendix Table~\ref{tab:coax_xai_schema}).
Interpreting \textit{Example-based XAI} (e.g., counterfactuals~\cite{mothilal2020explaining}) can be modeled by treating the example as an additional instance and leveraging the None reasoning strategy, or
defining a new reasoning strategy with attention on the changed attributes of counterfactual examples.
Interpreting 
\textit{Concept-based XAI} (e.g., TCAV~\cite{kim2018interpretability}, concept bottleneck~\cite{koh2020concept}) can be modeled by normalizing concept values and treating concepts as analogous to tabular attributes.

However, other XAI types would need further modifications to CoAX.
Interpreting \textit{Rules} (e.g., decision sets~\cite{lakkaraju2016interpretable}, Anchors~\cite{ribeiro2018anchors}) could be modeled by encoding threshold relations in memory chunks.
\textit{Partial dependence plots} (e.g., GAM~\cite{caruana2015intelligible}, ICE plots~\cite{goldstein2015peeking}) could be modeled with piecewise linear chunks to approximate 
\fixed{nonlinear relations}.

\subsubsection{Visualization Representation and Perception}

Many XAI methods present explanations as visualizations, such as tornado plots~\cite{ribeiro2016should}, lollipop charts~\cite{abdul2020cogam}, or SHAP scatter plots~\cite{lundberg2017unified}. 
We focused on user reasoning over the semantic content of Attribution and Importance XAI by parsing them before ingesting into CoAX. 
However, visual representations can influence or bias how attribution and importance are perceived and interpreted.
Particularly, poorly designed XAI visual representations can have low legibility, readability, and thus understandability~\cite{labarta2025cue}.
For our user studies, we had applied UX principles to iteratively design the UI with pilot feedback to improve legibility and interpretability.
We adapted an effective tabular UI showing key elements of feature attribution XAI~\cite{bo2024incremental, poursabzi2021manipulating} by including meter bars to indicate relative attribute values, excluding numeric labels for attribution subscores to reduce information overload (and their lack of use), etc.
Nevertheless, our design \fixed{choices}
may have inadvertently prioritized or suppressed certain strategies, and this should be studied.
Future work could extend CoAX by modeling visualization perception as an intermediate stage (e.g., square root attention of surface size~\cite{nixon1924attention}, visual attention in chart-reading~\cite{lohse1997models, shi2025chartist}), or adding cognitive parameters for UI readability and perceptual noise.

\subsubsection{Multimodal Explanations}

Beyond tabular data, \textit{Image-based XAI} (e.g., saliency maps~\cite{selvaraju2017grad}, segmentation~\cite{kirillov2023segany}) requires perceiving and mapping visual masks to semantic concepts before symbolic reasoning.
While CoAX addresses the latter, it can be extended with a concept extractor, implemented via object detection, segmentation, or learned visual concepts—that maps images to a set of interpretable features.
%
Similarly, AI is increasingly providing \textit{Natural Language XAI}, which require users to extract symbolic relations to build mental models of the model's logic~\cite{johnson1983mental}.
Future extensions of CoAX could simulate this process by integrating production-rule parsing (e.g., ACT-R style~\cite{lewis2005activation}) or leveraging LLMs to map textual explanations into structured attributes~\cite{tian2024large}.

\subsubsection{Decision and Real-World Tasks}
\label{sec:discussion_realworld}

We focused on forward simulation, a common foundational proxy task in XAI for testing understanding of the AI decision boundary.
 However, there remains a gap between user simulating AI outputs and improving real-world decisions, such as reliance, trust, and feature auditing.
\textit{AI reliance}  can be estimated by comparing the user simulated answer (from real users or CoAX) of the ground-truth (instead of the AI prediction), and calculating ratio metrics RAIR/RSR~\cite{schemmer2022should}. This leverages current practice in user studies on AI reliance~\cite{bo2025rely, he2023knowing, swaroop2025personalising}.
Accounting for how user preferences for reasoning strategies may shift with this downstream task~\cite{payne1993adaptive}, 
\textit{Trust} could be modeled by representing the user's mental model of two target labels---the AI decision and the ground-truth, and integrating them using decision-theoretic methods for user confidence~\cite{wang2022will} and information signals~\cite{guo2024decision}.
\textit{Fair feature auditing}~\cite{dodge2019explaining} can be modeled by 
encoding the user's social values as priority weights (cognitive parameters) and mapping to attribute sensitivity to discount or override decisions.
Empirically validating these task-specific strategy shifts remains a necessary direction for future work.

\section{Conclusion}

Despite significant advancements in Explainable AI (XAI), many techniques remain untested with real users or have been shown to be ineffective to improve user understanding. 
This work proposes an initial step with CoAX to overcome this hurdle by developing a virtual human proxy using cognitive models to simulate human cognition to support debugging flawed human reasoning strategies to interpret XAI. 
This will facilitate early testing of XAI, iterative prototyping of XAI, and reduce the cost of eventual user testing with real human participants. 
With the improved ease of testing, this can provide a basis for benchmark evaluation of XAI to support emerging AI-related regulatory compliance.



\begin{acks}
We thank Ee Wei Seah for her invaluable help in early discussions and initial development of the explanation user interface.
This research is supported by the National Research Foundation, Singapore and Infocomm Media Development Authority under its Trust Tech Funding Initiative (Award No. DTC-RGC-09).
Any opinions, findings and conclusions or recommendations expressed in this material are those of the author(s) and do not reflect the views of National Research Foundation, Singapore and Infocomm Media Development Authority.
\end{acks}

\section*{GenAI Usage Disclosure}   
The authors used generative AI tools (ChatGPT, Gemini) to assist with language refinement and clarity, and Table formatting.
The authors reviewed and edited all AI-generated content and take full responsibility for the final version of the manuscript.

\section*{Ethical Considerations Statement}
The summative experiment conducted prioritized participant privacy, excluding the collection of personal information. Each participant was assigned a
unique ID for identification, ensuring anonymity. Participants received detailed information about data collection and processing
and gave explicit consent to participate. They were also informed
of their right to withdraw at any point, which would lead to the
deletion of their data. The study was approved by the university institutional review board.

\bibliographystyle{ACM-Reference-Format}
\bibliography{manuscript}

\appendix

\section{Datasets, AI and XAI models}

\begin{table*}[h!]
\caption{Summary of datasets, AI models, explanation methods, attributes used (categorical/numeric), and predictive accuracy.}
\centering
\resizebox{\linewidth}{!}{
\begin{tabular}{lllll}
\toprule
\textbf{Dataset} & \textbf{Attributes (Type)} & \textbf{AI Model} & \textbf{XAI Method} & \textbf{Accuracy} \\
\midrule
\textbf{Forest Cover Type} &
\begin{tabular}[c]{@{}l@{}}
Elevation (num), Angle (num), \\
Dist to Water (num), \\
Dist to Road (num), \\
Hillshade (num)
\end{tabular} &
XGBoost (lr=0.05) &
SHAP &
0.877 \\
\midrule
\textbf{Adult Income} &
\begin{tabular}[c]{@{}l@{}}
Age (num), Years of Education (num), \\
Married (cat), Sex (cat), \\
Capital Gain (num)
\end{tabular} &
XGBoost (lr=0.05) &
LIME &
0.8133 \\
\midrule
\textbf{Wine Quality} &
\begin{tabular}[c]{@{}l@{}}
Vinegar Taint (num), SO2 (num), pH (num), \\
Sulphates (num), Alcohol (num)
\end{tabular} &
MLP (2 layers, 50 units, lr=$10^{-3}$) &
LIME &
0.7917 \\
\bottomrule
\end{tabular}
}
\label{tab:datasets-models}
\end{table*}

\section{Experimental Design and Procedure}

\begin{figure}[htbp!]
  \centering
  \includegraphics[width=0.95\textwidth]{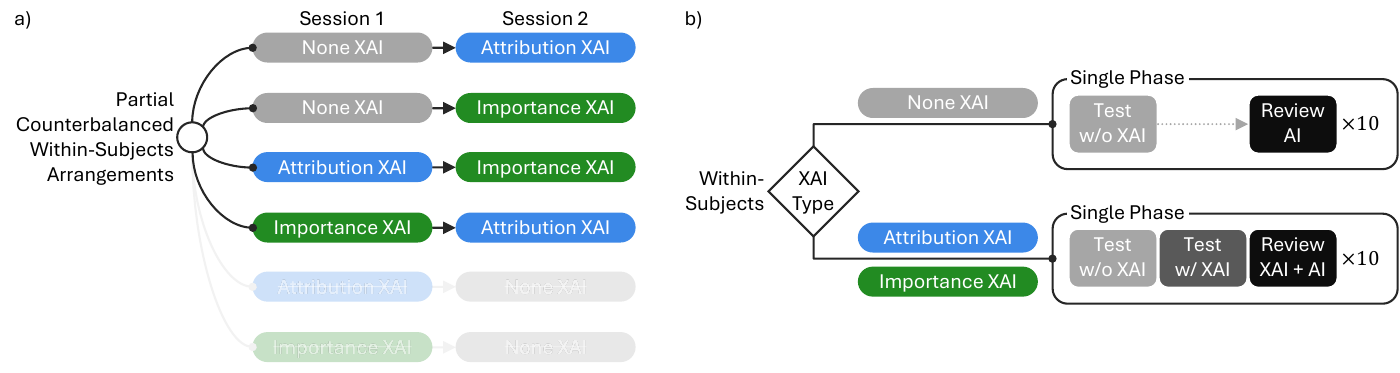}  
    \caption{Experiment arrangements (a) and main section pipeline (b) in formative user study for 2-session partial counterbalancing of XAI Types in within-subjects experiment design. Faded out omitted arrangements with None XAI in Session 2 to avoid learning effects due to earlier XAI in Session 1.}
    \label{fig:formative-experimental-design}
\end{figure}



\clearpage
\section{Tukey HSD results by XAI type}
\label{sec:appendix-tukey-hsd-by-xaitype}

We show the Tukey test results across XAI types and testing with XAI conditions

\begin{table}[htbp!]
\centering
\caption{Tukey HSD test at $\alpha = 0.05$ of Human responses and CoAX simulation for the forward simulation task on the Adult Income dataset across XAI Type and with or without XAI testing}
\label{tab:human-coax-comparison}
\begin{tabular}{lcccccccccc}
\toprule
\textbf{Level} 
& \multicolumn{4}{c}{\textbf{Human}} 
& \multicolumn{5}{c}{\textbf{CoAX}} \\
\cmidrule(lr){2-5} \cmidrule(lr){6-10}
& A & B & C & Mean & A & B & C & D & Mean \\
\midrule
Attribution: w/ XAI   
& A &   &   & 0.858 
& A &   &   &   & 0.789 \\

Importance: w/ XAI   
& A &   &   & 0.801 
& A & B &   &   & 0.774 \\

Attribution: w/o XAI 
& A & B &   & 0.797 
&   & B & C &   & 0.734 \\

Importance: w/o XAI  
&   & B &   & 0.737 
&   &   & C &   & 0.693 \\

None: w/o XAI        
&   & B &   & 0.684 
&   &   &   & D & 0.605 \\

\bottomrule
\end{tabular}
\end{table}

\begin{table}[htbp!]
\centering
\caption{Tukey HSD test at $\alpha = 0.05$ of Human responses and CoAX simulation for the forward simulation task on the Forest Cover dataset across XAI Type and with or without XAI testing}
\label{tab:human-coax-comparison-2}
\begin{tabular}{lcccccccccc}
\toprule
\textbf{Level} 
& \multicolumn{4}{c}{\textbf{Human}} 
& \multicolumn{5}{c}{\textbf{CoAX}} \\
\cmidrule(lr){2-5} \cmidrule(lr){6-10}
& A & B & C & Mean & A & B & C & D & Mean \\
\midrule
Attribution: w/ XAI   
& A &   &   & 0.947 
& A &   &   &   & 0.964 \\

Importance: w/ XAI   
&   & B &   & 0.617 
&   & B &   &   & 0.646 \\

Attribution: w/o XAI 
&   & B &   & 0.566 
&   &   & C &   & 0.562 \\

Importance: w/o XAI  
&   & B &   & 0.586 
&   &   & C &   & 0.575 \\

None: w/o XAI        
&   & B &   & 0.580 
&   &   & C &   & 0.558 \\

\bottomrule
\end{tabular}
\end{table}

\begin{table}[htbp!]
\centering
\caption{Tukey HSD test at $\alpha = 0.05$ of Human responses and CoAX simulation for the forward simulation task on the Wine Quality dataset across XAI Type and with or without XAI testing}
\label{tab:human-coax-comparison-3}
\begin{tabular}{lcccccccccc}
\toprule
\textbf{Level} 
& \multicolumn{4}{c}{\textbf{Human}} 
& \multicolumn{5}{c}{\textbf{CoAX}} \\
\cmidrule(lr){2-5} \cmidrule(lr){6-10}
& A & B & C & Mean & A & B & C & D & Mean \\
\midrule
Attribution: w/ XAI   
& A &   &   & 0.838 
& A &   &   &   & 0.809 \\

Importance: w/ XAI   
&   & B & C & 0.654 
&   & B &   &   & 0.658 \\

Attribution: w/o XAI 
&   & B & C & 0.667 
&   & B &   &   & 0.682 \\

Importance: w/o XAI  
&   & B &   & 0.691 
&   & B &   &   & 0.688 \\

None: w/o XAI        
&   &   & C & 0.613 
&   & B &   &   & 0.648 \\

\bottomrule
\end{tabular}
\end{table}

\clearpage
\section{Hyperparameter Tuning}
\label{sec:appendix-tuning-hyperparams}
We show the hyperparameters tuned for the CoAX models as well as the baseline ML proxies, as well as the distribution of the fitted parameters for the CoAX model

\subsection{Hyperparameter Tuning Ranges}
\label{sec:appendix-hyperparam-ranges}

Refer to Table \ref{tab:hyperparameter-space} for the list of tuned hyperparameters and their values

\begin{table*}[htbp!]
\centering
\caption{Hyperparameter search space for cognitive strategies and machine learning proxy models. Parameters shared across cognitive strategies (e.g., $\alpha$, $\rho$, $k$) are grouped. Attribution sum uses $\zeta$ instead of $\alpha$.}
\label{tab:hyperparameter-space}
\begin{tabular}{llcl}
\toprule
Parameter & Strategies & Type & Range / Description \\
\midrule
Exemplar sensitivity $\alpha$ & 
\begin{tabular}[c]{@{}l@{}}
Sensitive-features categorization \\
Salient-features categorization \\
Importance categorization
\end{tabular} & 
Real & [1.0, 40.0] \\

Number of features $k$ & All cognitive strategies & Integer & [1, 4] \\

Retrieval threshold $\rho$ & All cognitive strategies & Real & [-2.8, -1.5] \\

Scaling factor $\zeta$ & Attribution sum & Real & [0.1, 5.0] \\

Memory decay rate $\lambda$ & All cognitive strategies & Constant & 0.5 (fixed) \\
\midrule
Max depth & DecisionTree & Integer & [1, 5] \\
Num neighbors & KNN & Integer & [1, 8] \\
Hidden dimension & MLP & Integer & [1, 50] \\
Smoothing factor & All ML models & Real & [0.0, 5.0] \\
\bottomrule
\end{tabular}
\end{table*}

\subsection{Fitted Hyperparameters}

\begin{table}[h]
\centering
\caption{Average fitted hyperparameter means and standard deviations for each cognitive strategy}
\resizebox{\linewidth}{!}{
\begin{tabular}{lcccc}
\toprule
\textbf{Strategy} 
& \textbf{Exemplar sensitivity $\alpha$} 
& \textbf{Number of features $k$} 
& \textbf{Retrieval threshold $\rho$} 
& \textbf{Scaling factor $\zeta$} \\
\midrule
Attribution sum 
& -- 
& $3.750 \pm 1.073$ 
& $-2.567 \pm 1.149$ 
& $2.828 \pm 2.356$ \\

Importance categorization 
& $31.365 \pm 35.320$ 
& $2.772 \pm 1.661$ 
& $-2.284 \pm 1.508$ 
& -- \\

Salient-features categorization 
& $28.608 \pm 35.173$ 
& $2.253 \pm 1.160$ 
& $-2.624 \pm 1.447$ 
& -- \\

Sensitive-features categorization 
& $24.221 \pm 31.737$ 
& $2.948 \pm 1.578$ 
& $-2.337 \pm 1.304$ 
& -- \\
\bottomrule
\end{tabular}}
\label{tab:avg-mu-sd-hyperparams-by-strategy}
\end{table}

\clearpage
\section{Prevalence and Accuracy Across Datasets}
\label{sec:appendix-prevalence-strategy}
In the main section, we visualized the matching of the CoAX model to the human behavior only for the Adult Income dataset. Here, we provide similar visualizations for the other 2 datasets.

\begin{figure}[h!]
  \centering
  \includegraphics[width=0.82\textwidth]{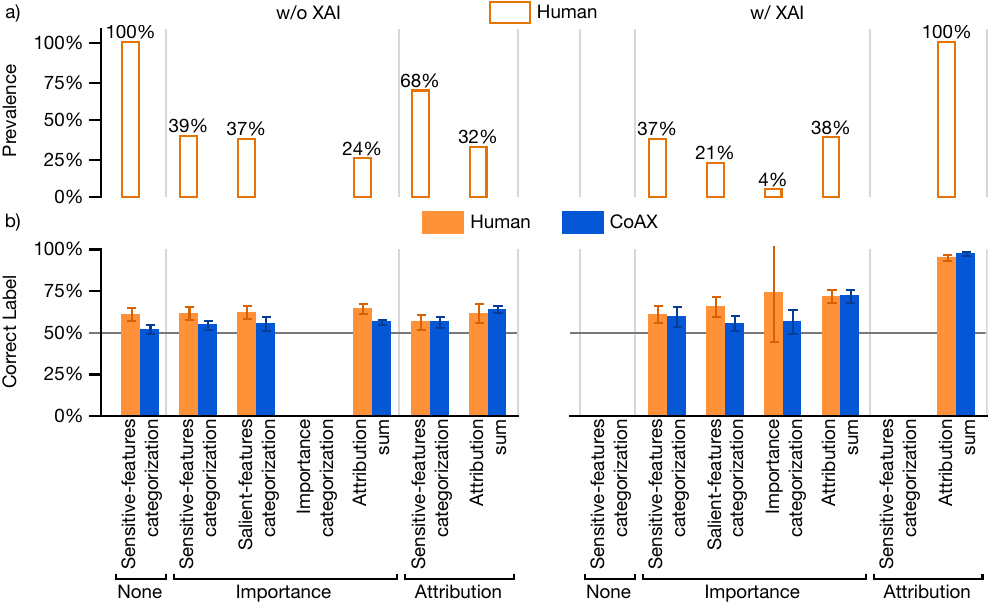}
    \caption{
    Results of summative user study (orange) and CoAX simulation (blue) of prevalence (a) and label correctness (b) by Reasoning Strategy for the Forest Cover Type dataset.
    Error bars indicate 95\% CI. 
    }
    \label{fig: covertype-strategy-comparison}
\end{figure}

\begin{figure}[h!]
  \centering
  \includegraphics[width=0.82\textwidth]{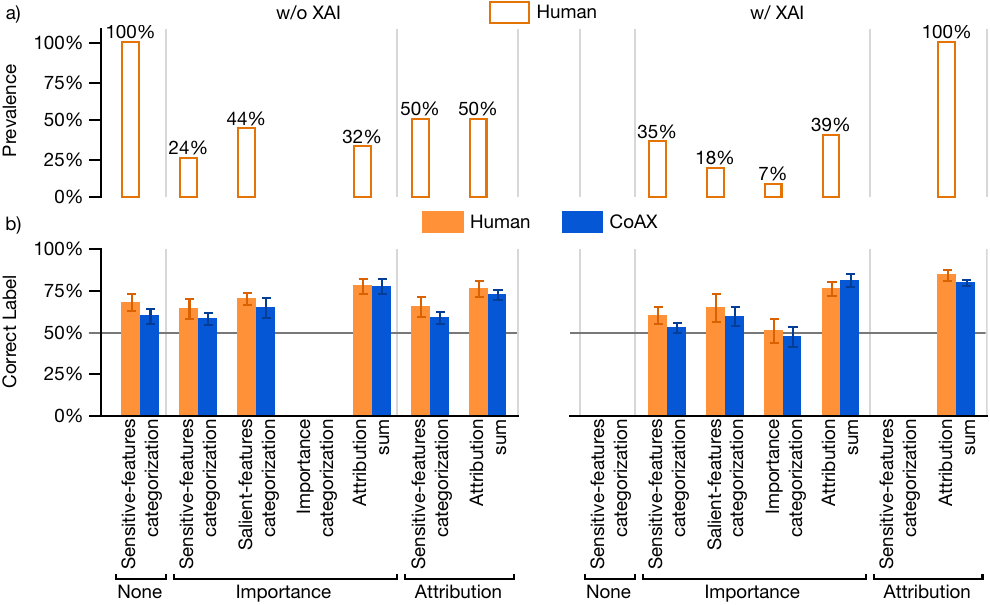}
    \caption{
    Results of summative user study (orange) and CoAX simulation (blue) of prevalence (a) and label correctness (b) by Reasoning Strategy for the Wine Quality dataset. 
    Error bars indicate 95\% CI.
    }
    \label{fig: wine-quality-strategy-comparison}
\end{figure}



\clearpage
\section{Statistical Results from Tukey HSD Tests}
\label{sec:appendix-tukey-hsd-strategies}


\begin{table}[h]
  \centering
  \caption{
    Tukey HSD test at $\alpha = 0.05$ of Human responses and CoAX simulation for the forward simulation task on the Forest Cover Type dataset across XAI Type, with or without XAI testing and strategy.
  }
  \includegraphics[width=12.4cm]{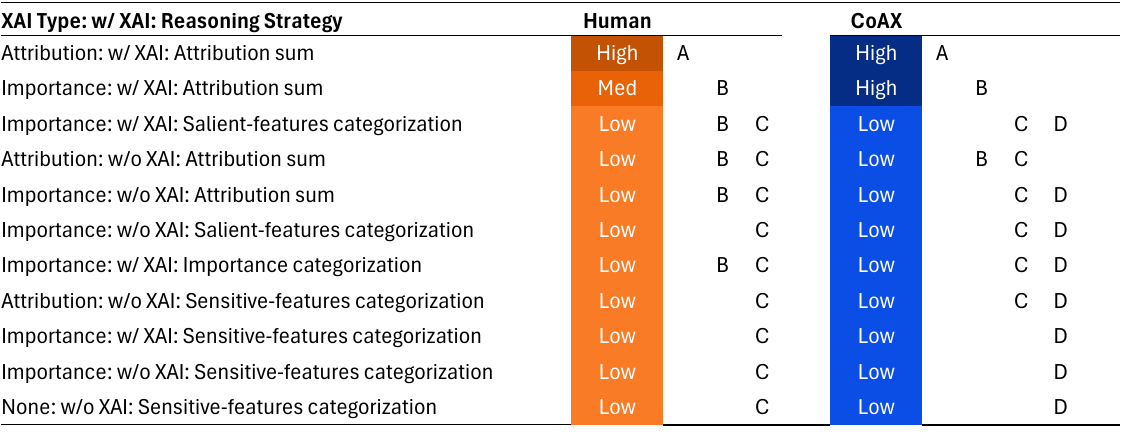}
  \label{tab:tukey-hsd-forest-cover-type-strategies}
\end{table}

\begin{table}[h]
  \centering
  \caption{
    Tukey HSD test at $\alpha = 0.05$ of Human responses and CoAX simulation for the forward simulation task on the Wine Quality dataset.
  }
  \includegraphics[width=12.4cm]{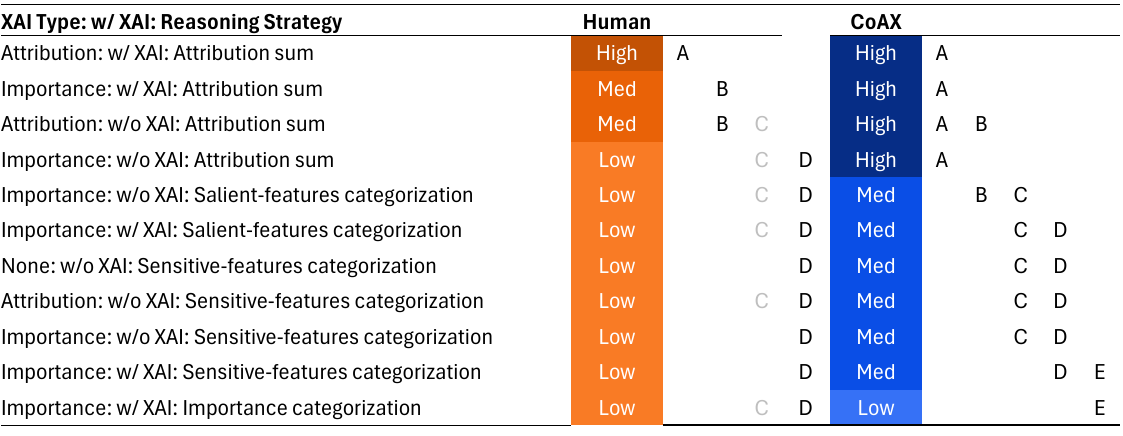}
  \label{tab:tukey-hsd-wine-quality-strategies}
\end{table}

\newpage
\section{CoAX Against Baseline Machine Learning Proxies on Other Datasets}
\label{sec:appendix-coax-vs-baselines}

In the main section, we visualized the fitting results to real participants' responses of CoAX against other baseline ML proxies only for the Adult Income dataset. Here, we provide similar visualizations for the other 2 datasets.


\begin{figure}[h!]
  \centering
  \includegraphics[width=1.0\textwidth]{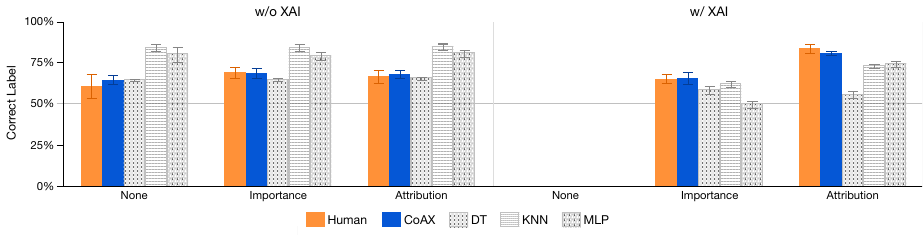}  
    \caption{
    Results of summative user study (orange) compared to virtual proxies (CoAX blue, ML-based grey) of label correctness by XAI Type for the Wine Quality dataset.
    Error bars 95\% CI.
    }
    \label{fig: wine-quality-baseline-comparison}
\end{figure}

\begin{figure}[h!]
  \centering
  \includegraphics[width=1.0\textwidth]{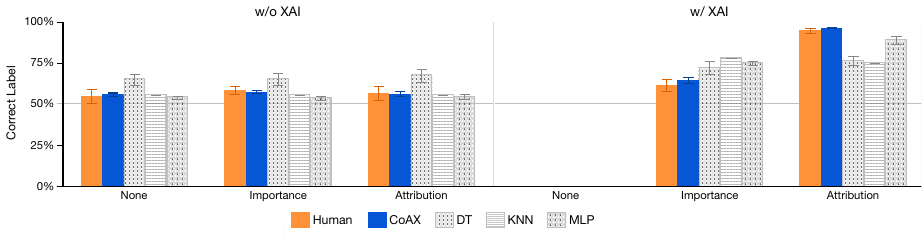}  
    \caption{
    Results of summative user study (orange) compared to virtual proxies (CoAX blue, ML-based grey) of label correctness by XAI Type for the Forest Cover Type dataset.
    Error bars 95\% CI.
    }
    \label{fig: covertype-baseline-comparison}
\end{figure}


\newpage
\section{Modeling Hypothesis}
\subsection{Experimental Variation: Number of Training Instances and Attributes}


\begin{figure}[h!]
  \centering
  \includegraphics[width=0.95\textwidth]{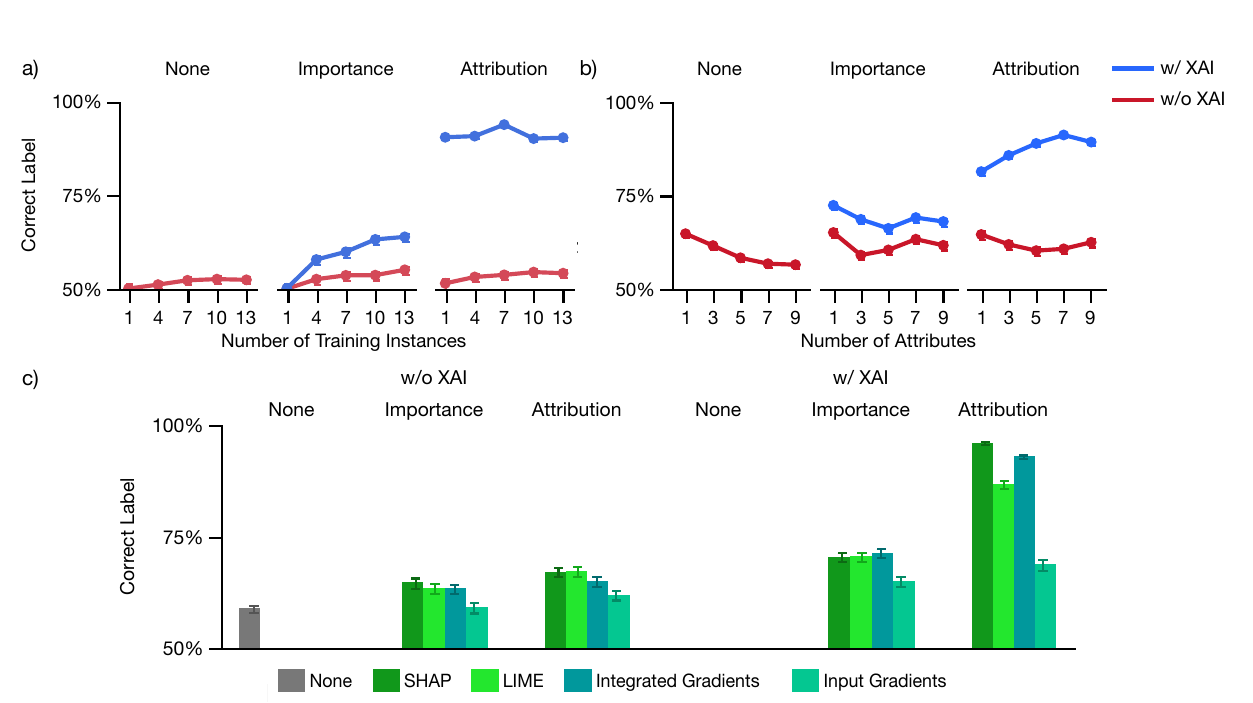}
  \caption{Modeling hypotheses study results of forward simulation correctness on the Forest Cover Type dataset, by XAI type, testing condition, a) number of training instances, b) number of attributes, and c) XAI model.}
  \label{fig:vary-num-training-and-attributes-forest-cover}
\end{figure}




\newpage
\section{Extension of CoAX}

\begin{table}[h]
\caption{Scope of CoAX for user interpretation of Local XAI Schemas on interpretable features.}
\label{tab:coax_xai_schema}
\centering
\begin{tabular}{p{0.18\linewidth} p{0.28\linewidth} p{0.50\linewidth}}
\hline
\textbf{XAI Schema} & \textbf{Representative XAI Methods} & \textbf{CoAX Coverage} \\
\hline
Attribution &
LIME~\cite{ribeiro2016should}, SHAP~\cite{lundberg2017unified}, Integrated Gradients~\cite{sundararajan2017axiomatic}, Input Gradients~\cite{shrikumar2017learning} &
Covered. \\

Example-based &
Prototypes~\cite{li2018deep}, Criticisms~\cite{kim2016examples}, Counterfactuals~\cite{mothilal2020explaining}, Semi-factuals~\cite{aryal2024semi} &
Direct extension: add as an additional exemplar to memory. More realistic approaches could involve additional strategies, e.g., focusing only on changed features with counterfactuals. \\

Concept-based &
TCAV~\cite{kim2018interpretability}, Concept Bottleneck Models (CBM)~\cite{koh2020concept} &
Direct extension: treat concepts as interpretable features, analogous to tabular attributes. \\

Rule-based &
Decision Sets~\cite{lakkaraju2016interpretable}, Anchors~\cite{ribeiro2018anchors} &
Future work: store rules instead of or in addition to exemplars in CoAX memory by encoding threshold relations in memory chunks. \\

Partial Dependence Plot &
Ceteris Paribus (CP) Plots~\cite{kuzba2019pyceterisparibus}, Individual Conditional Expectation (ICE)~\cite{goldstein2015peeking}, GAM~\cite{caruana2015intelligible} &
Future work: model with piecewise linear chunks encoded into memory to approximate nonlinear relations. \\

\hline
\end{tabular}
\end{table}

\newpage 
\section{Screenshots of Survey}
In this section, we provide screenshots of the quintessential parts of our survey. First the participant is introduced to the domain, then to the interface and asked screening questions, before proceeding to the main task of predicting the AI prediction. In the training phase, the participant will be given feedback on their responses.

\begin{figure}[h]
  \centering
  \includegraphics[width=0.6\textwidth]{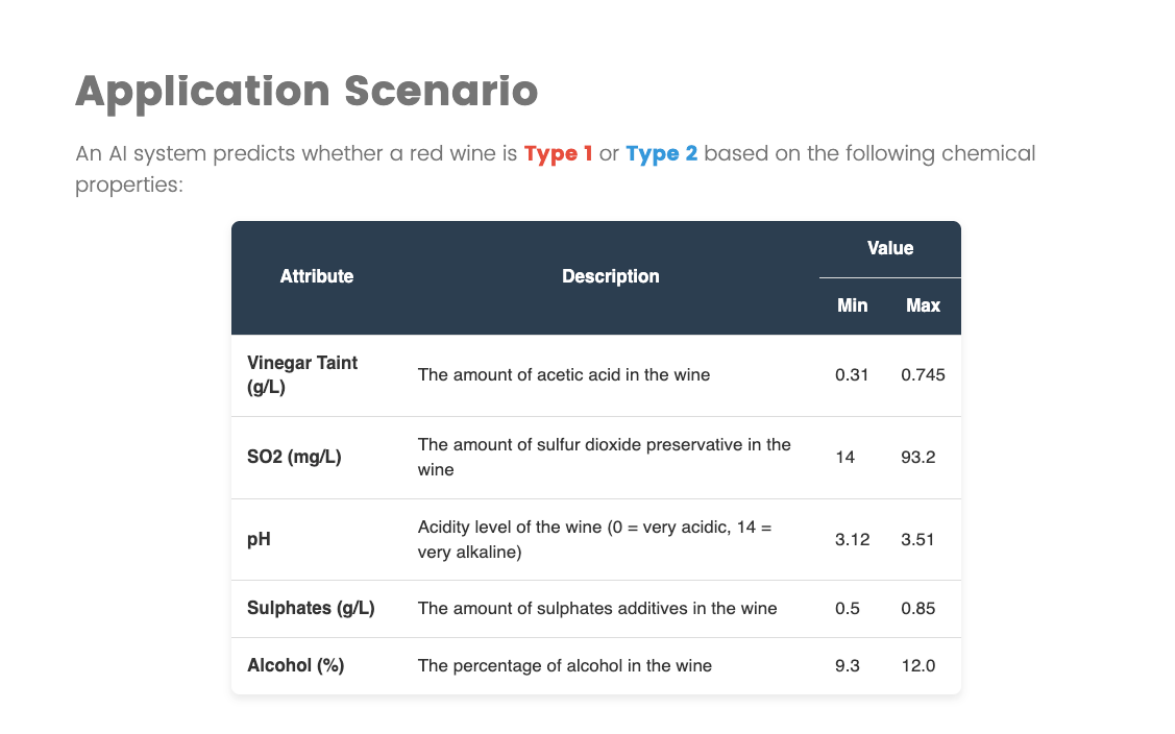}  
    \caption{
    Introduction to the application (dataset) domain.
    }
    \label{fig:survey-introduction}
\end{figure}

\begin{figure}[h]
  \centering
  \includegraphics[width=0.7\textwidth]{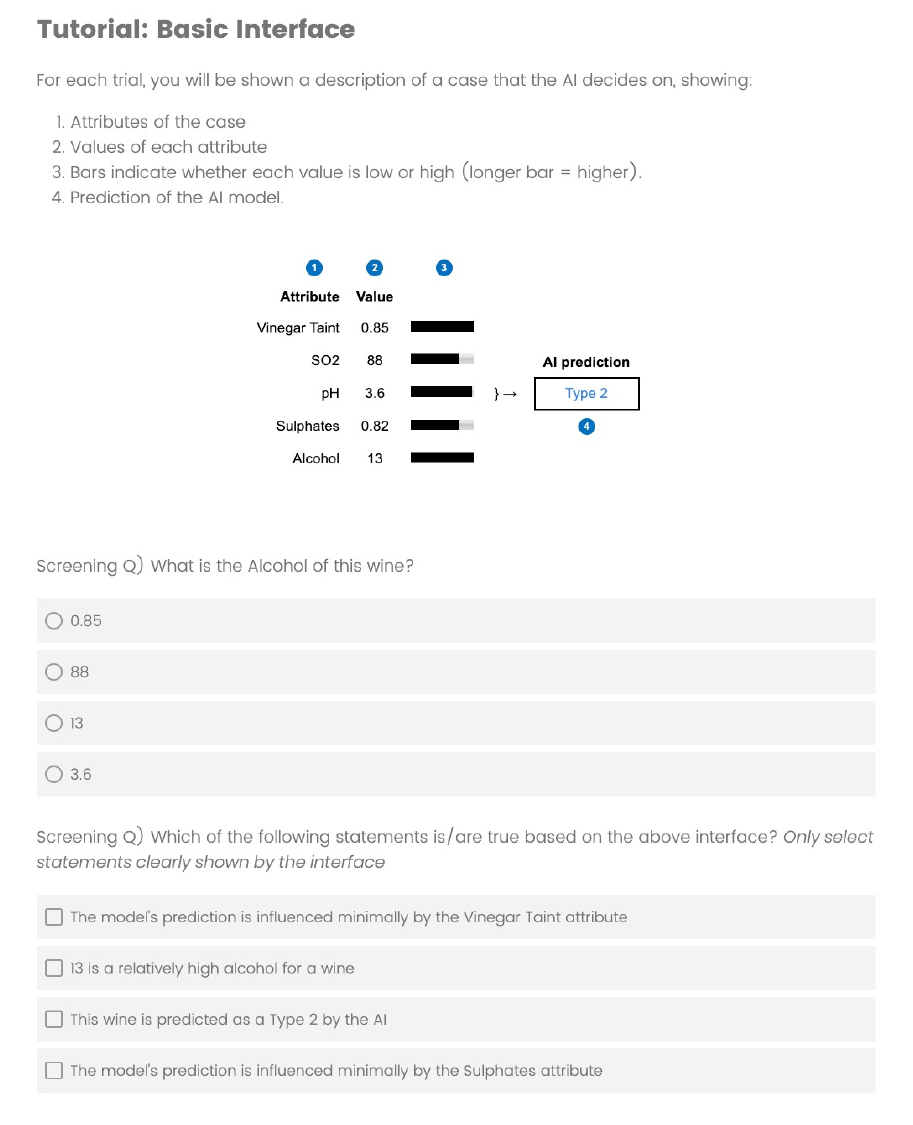}  
    \caption{
    Initial UI comprehension screening.
    }
    \label{fig:survey-initial-screening}
\end{figure}

\begin{figure}[h]
  \centering
  \includegraphics[width=0.7\textwidth]{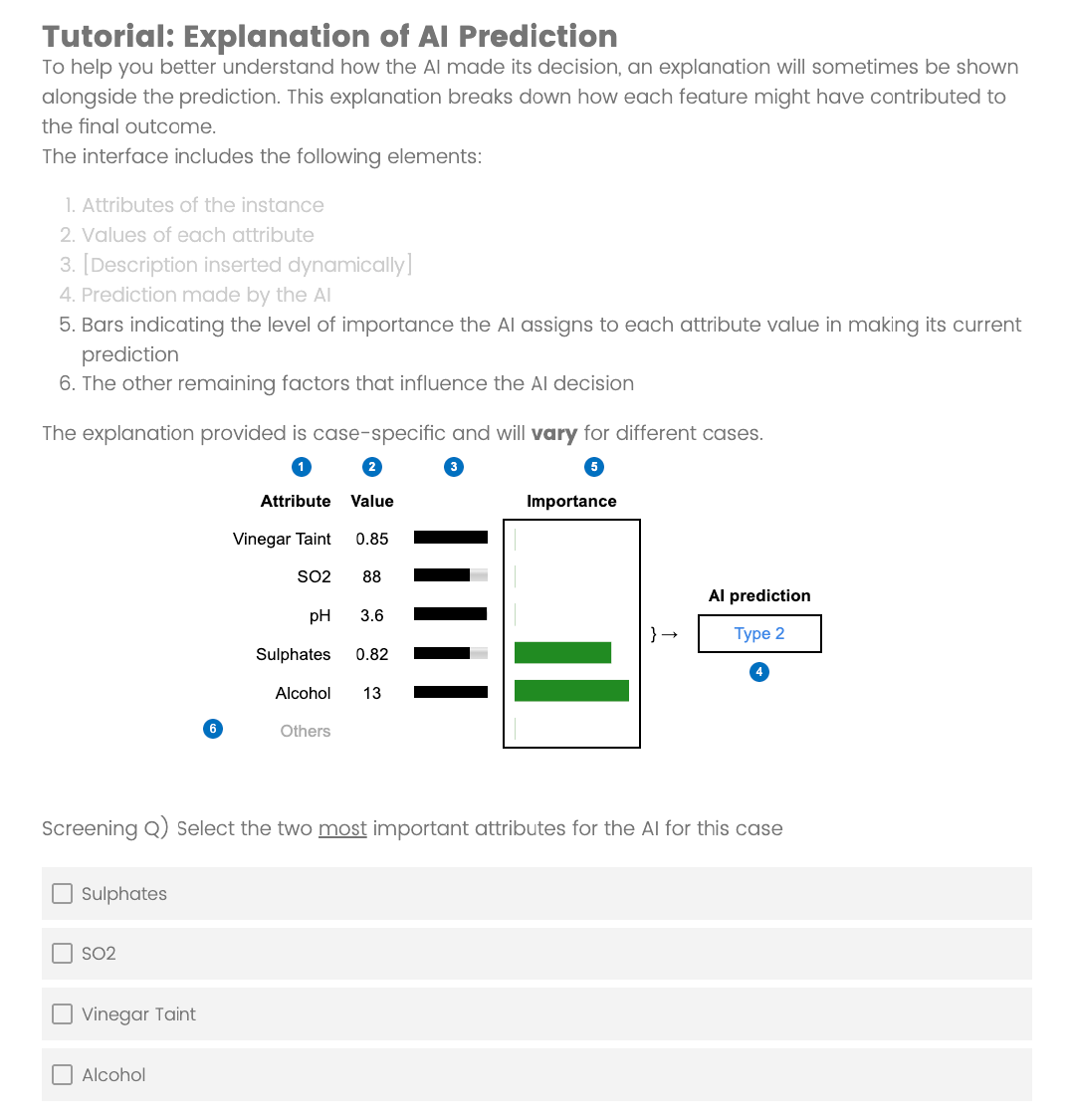}  
    \caption{
    Comprehension screening for the Importance explanation.
    }
    \label{fig:survey-screening-importancexai}
\end{figure}

\begin{figure}[h]
  \centering
  \includegraphics[width=0.7\textwidth]{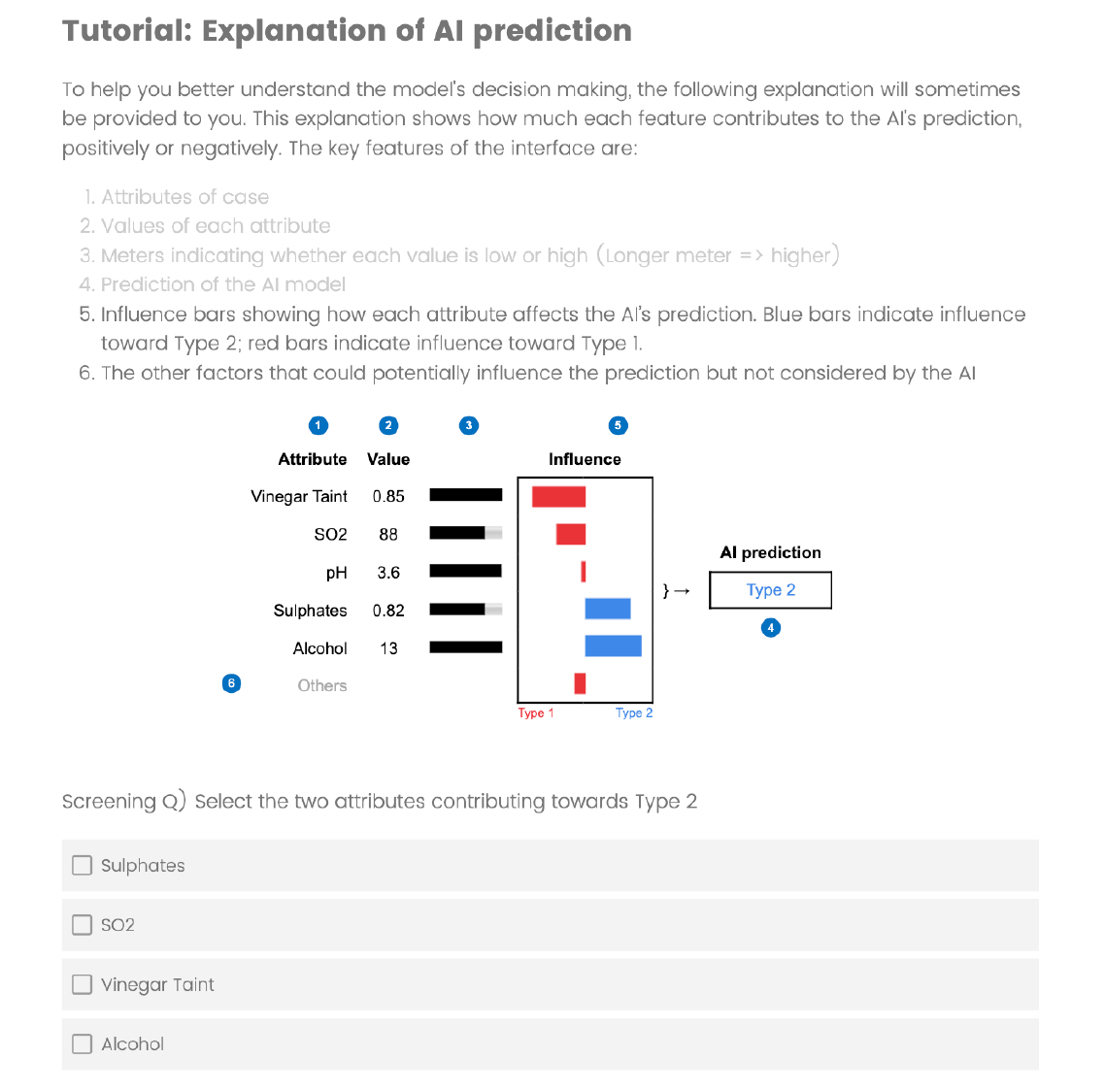}  
    \caption{
    Comprehension screening for the Attribution explanation.
    }
    \label{fig:survey-screening-attributionxai}
\end{figure}

\newpage
\begin{figure}[h]
  \centering
  \includegraphics[width=0.6\textwidth]{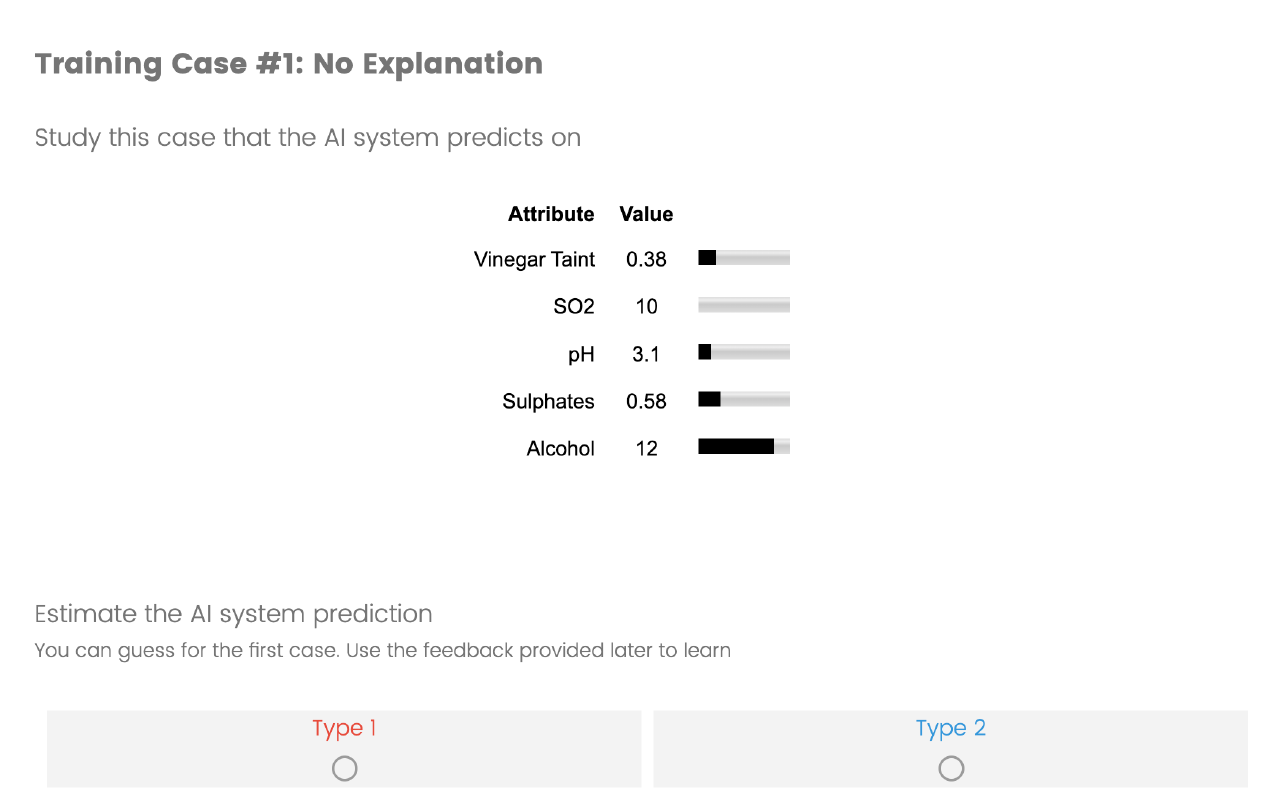}  
    \caption{
    Forward Simulation w/o XAI. Participant is required to provide binary choice.
    }
    \label{fig:survey-training-no-xai}
\end{figure}

\begin{figure}[h]
  \centering
  \includegraphics[width=0.6\textwidth]{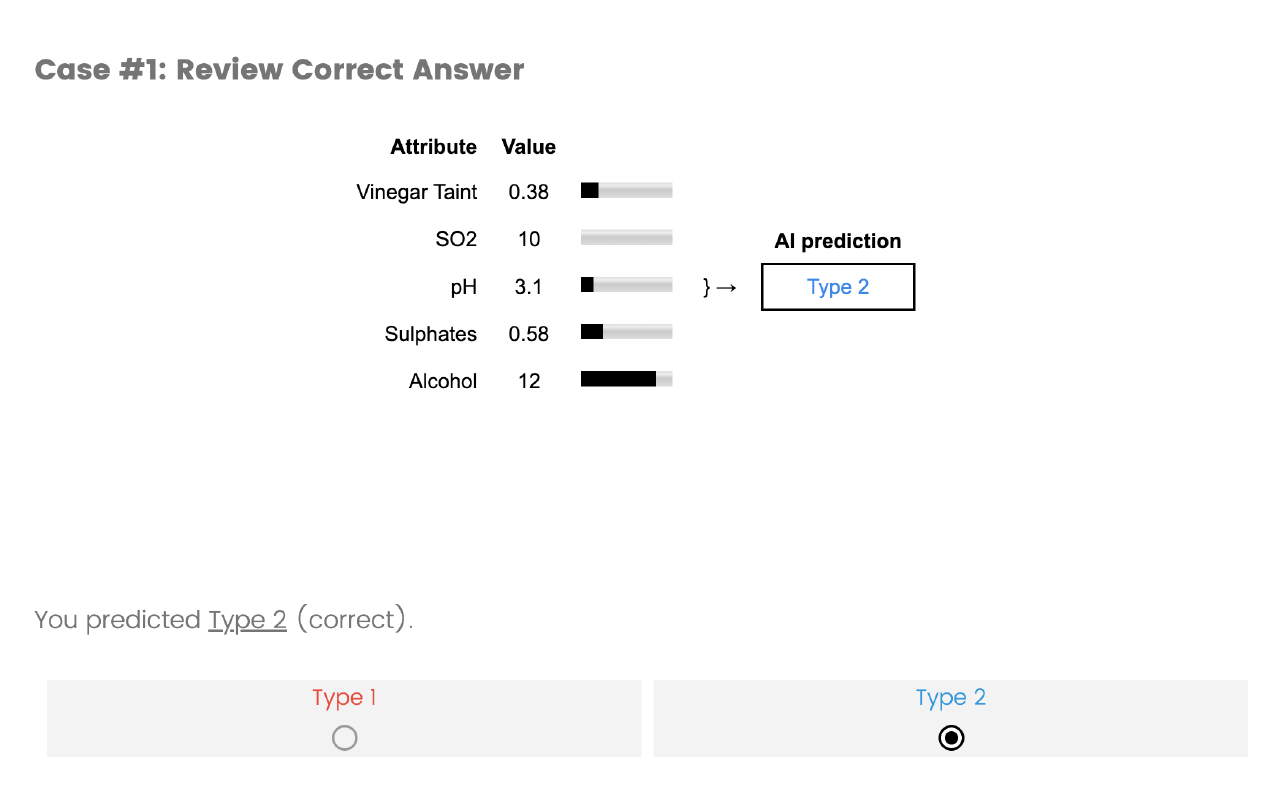}  
    \caption{
    Feedback on participant response (without XAI). 
    }
    \label{fig:survey-training-feedback-no-xai}
\end{figure}

\begin{figure}[h]
  \centering
  \includegraphics[width=0.6\textwidth]{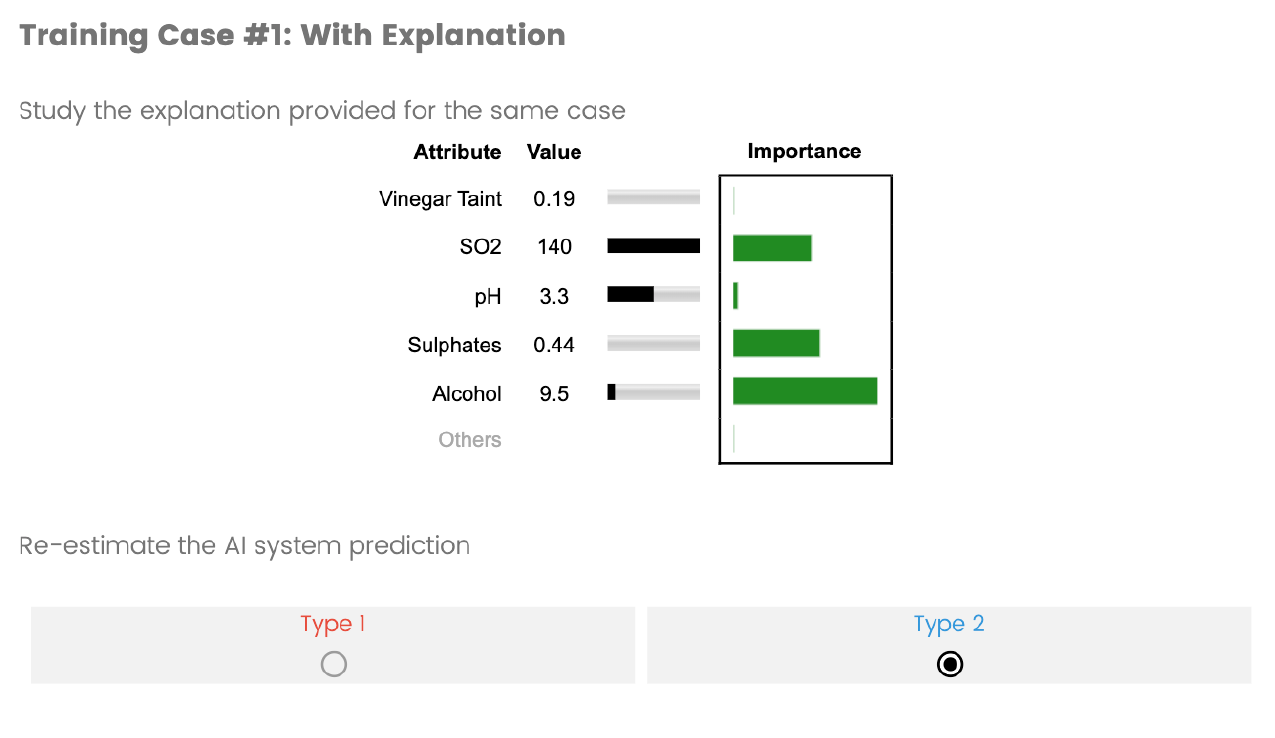}  
    \caption{
    Forward Simulation w/ Importance XAI. Participant is required to provide binary choice.
    }
    \label{fig:survey-training-w-importance-xai}
\end{figure}

\begin{figure}[h]
  \centering
  \includegraphics[width=0.6\textwidth]{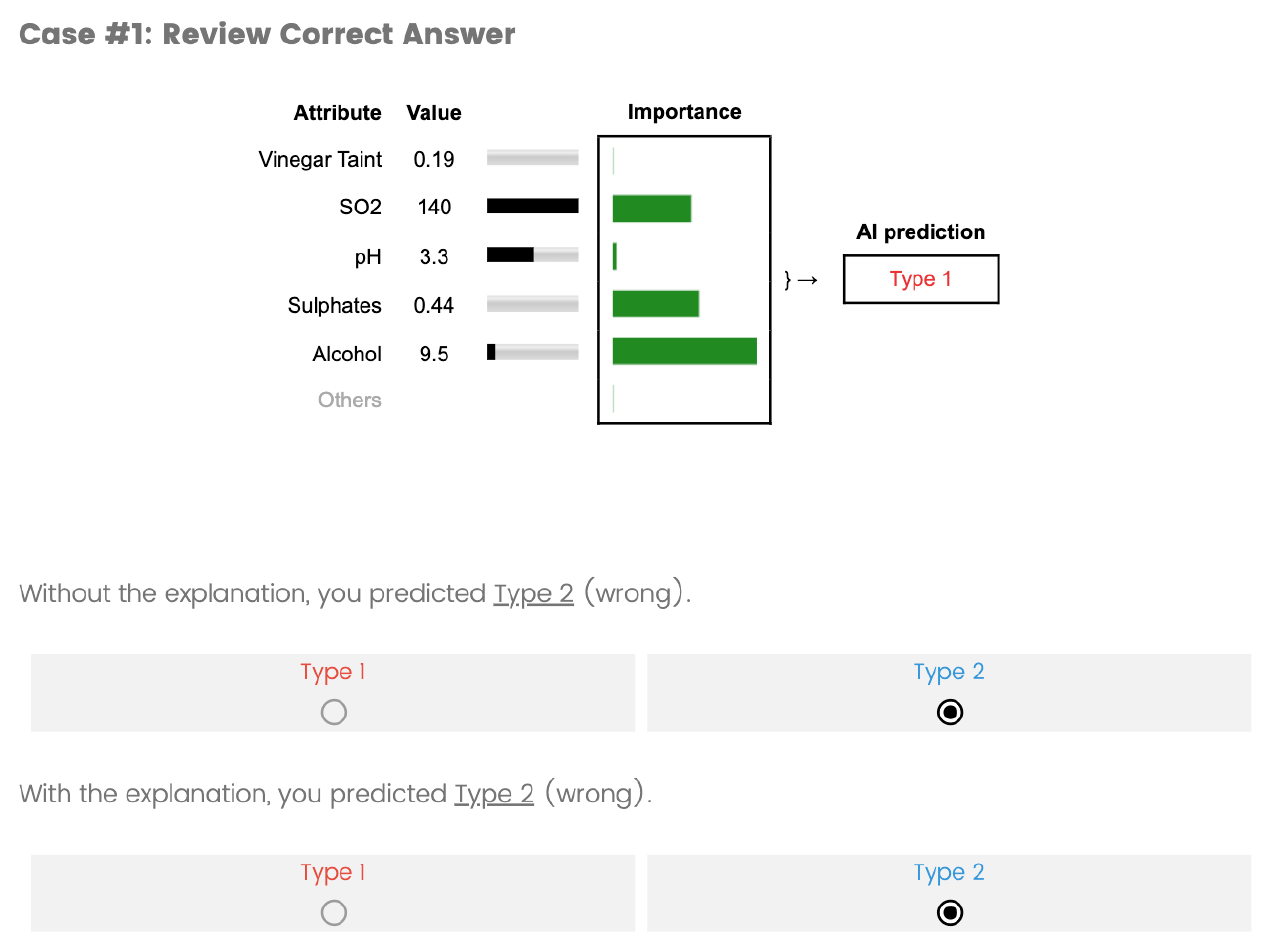}  
    \caption{
    Feedback on participant responses (with Importance XAI).
    }
    \label{fig:survey-training-feedback-w-importance-xai}
\end{figure}

\begin{figure}[h]
  \centering
  \includegraphics[width=0.6\textwidth]{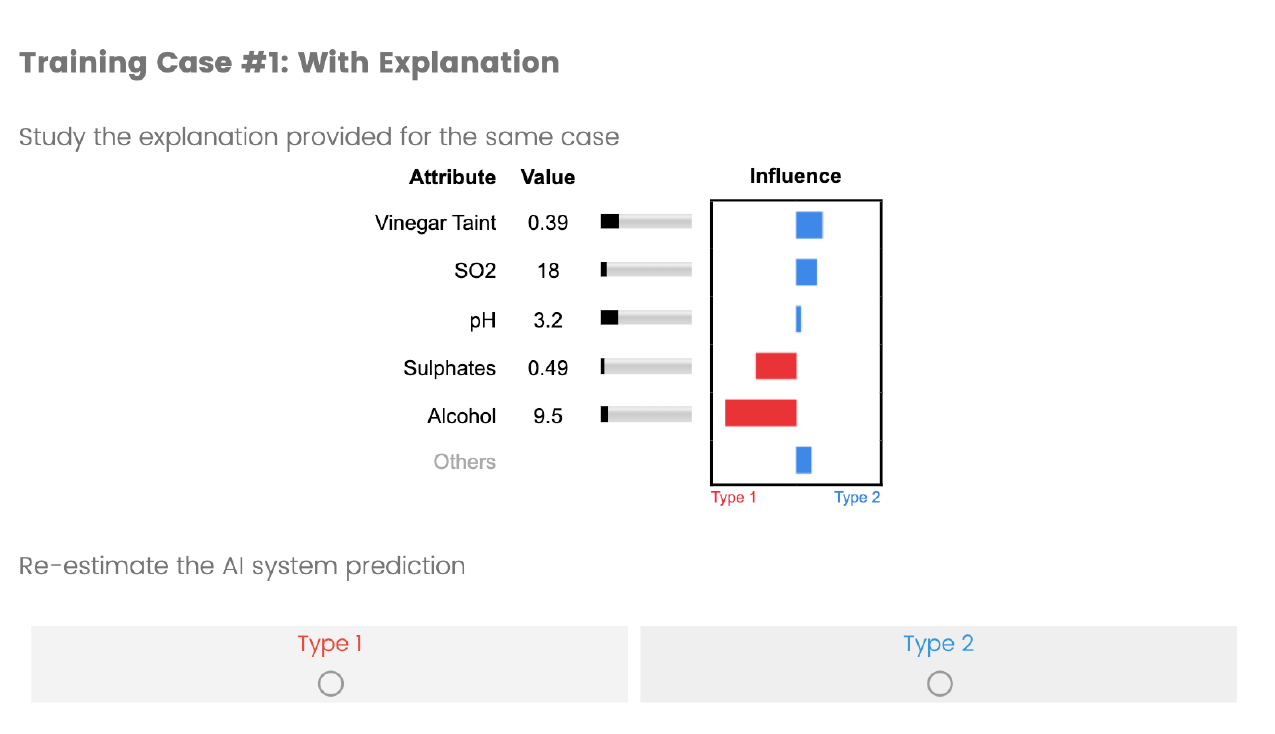}  
    \caption{
    Forward Simulation w/ Attribution XAI. Participant is required to provide binary choice.
    }
    \label{fig:survey-training-w-attribution-xai}
\end{figure}

\begin{figure}[h]
  \centering
  \includegraphics[width=0.6\textwidth]{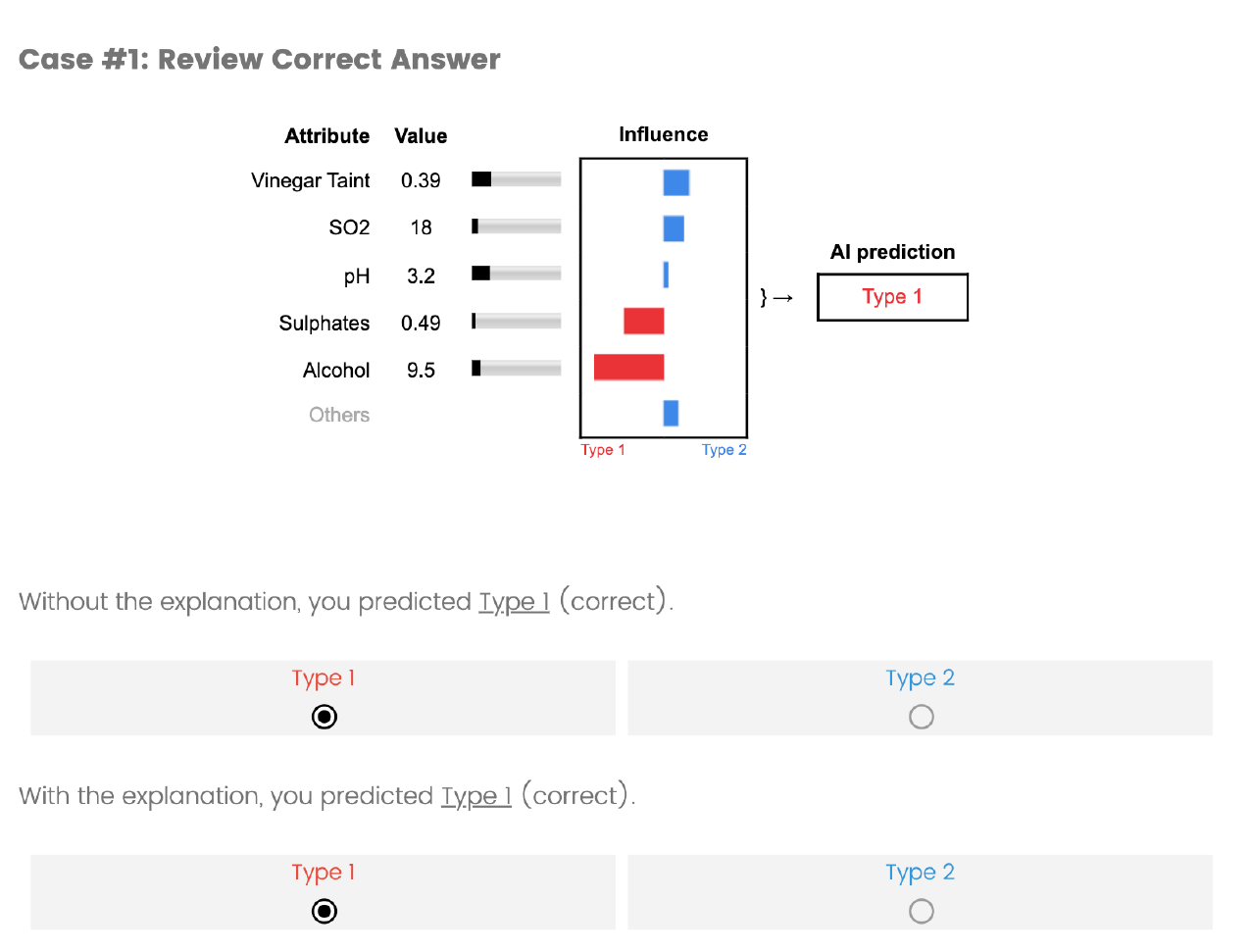}  
    \caption{
    Feedback on participant responses (with Attribution XAI).
    }
    \label{fig:survey-training-feedback-w-attribution-xai}
\end{figure}

\begin{figure}[h]
  \centering
  \includegraphics[width=0.6\textwidth]{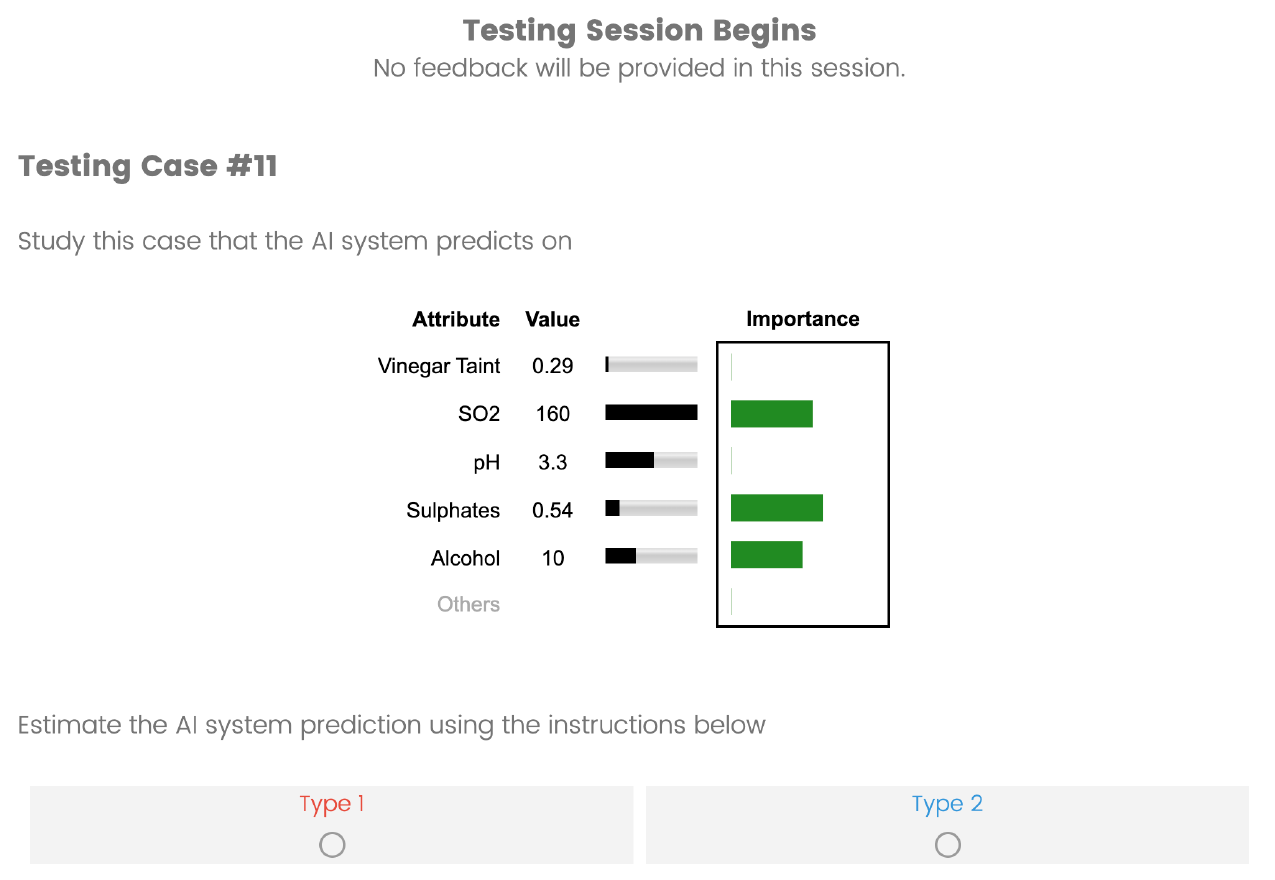}  
    \caption{
    Testing trial (with Importance XAI).
    }
    \label{fig:survey-testing-w-importance-xai}
\end{figure}

\begin{figure}[h]
  \centering
  \includegraphics[width=0.6\textwidth]{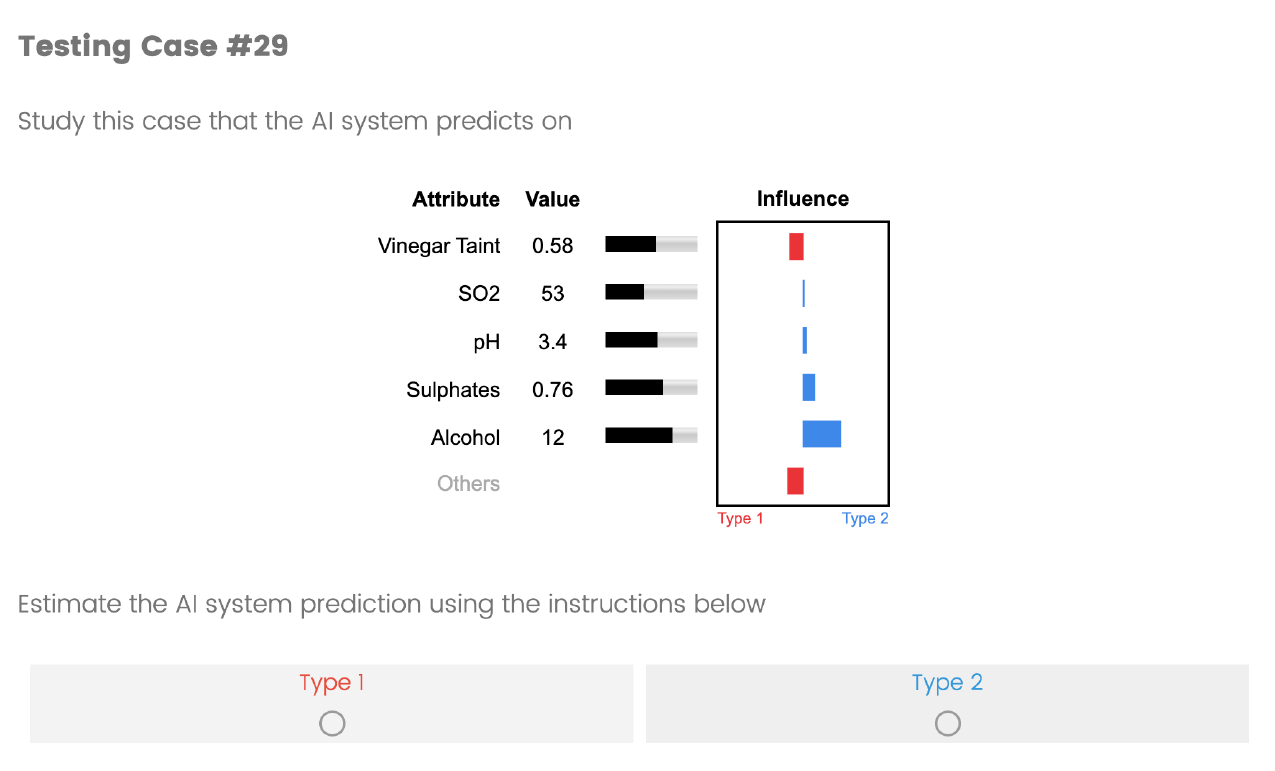}  
    \caption{
    Testing trial (with Attribution XAI).
    }
    \label{fig:survey-testing-w-attribution-xai}
\end{figure}

\begin{figure}[h]
  \centering
  \includegraphics[width=0.6\textwidth]{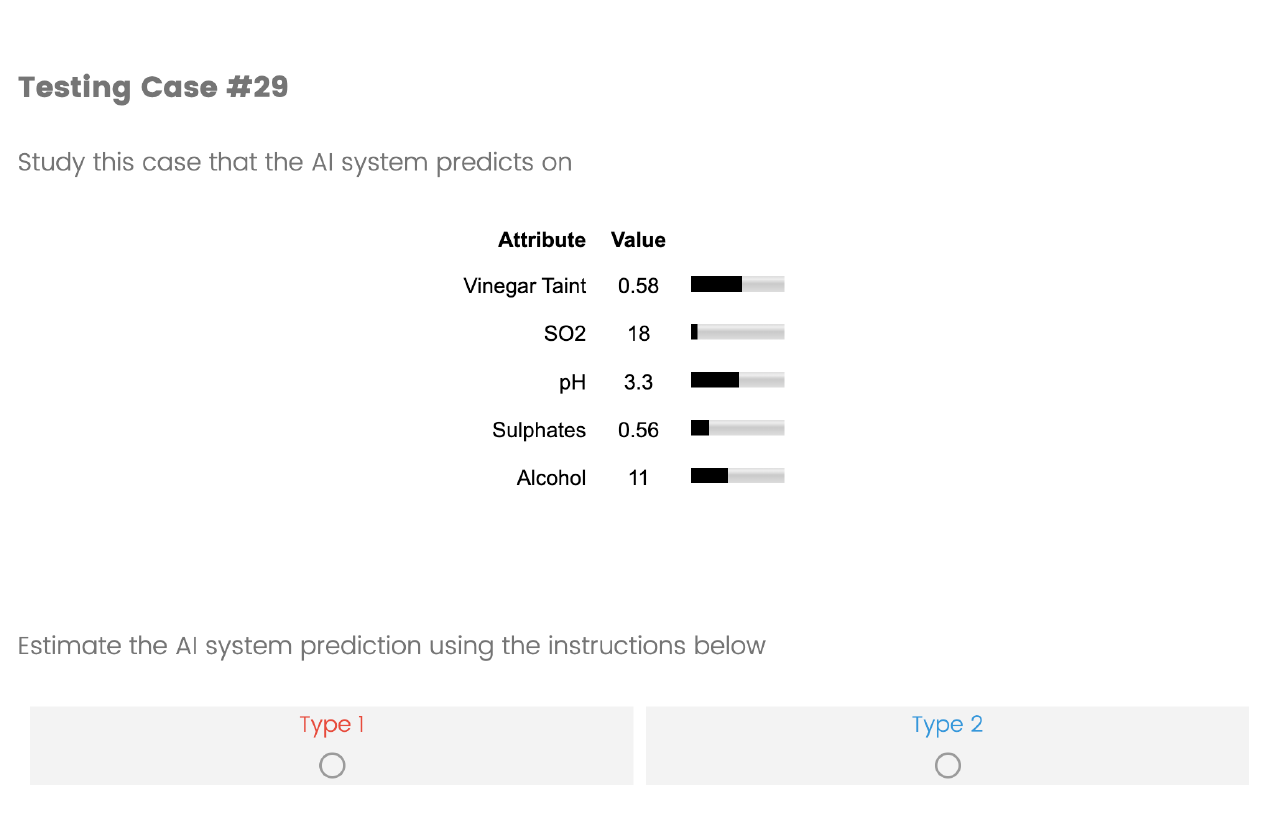}  
    \caption{
    Testing trial (without XAI).
    }
    \label{fig:survey-testing-w-no-xai}
\end{figure}
\end{document}